\documentclass[10pt,twocolumn,letterpaper]{article}
\usepackage{titling}

\usepackage[pagenumbers]{cvpr} % To force page numbers, e.g. for an arXiv version

\usepackage{graphicx}
\usepackage{amsmath}
\usepackage{amssymb}
\usepackage{booktabs}
\usepackage{multirow}

\usepackage[pagebackref,breaklinks,colorlinks]{hyperref}

\usepackage{color, colortbl}
\definecolor{LightGray}{rgb}{0.92,0.92,0.92}
\definecolor{Gray1}{rgb}{0.95,0.95,0.95}
\definecolor{Gray2}{rgb}{0.9,0.9,0.9}
\definecolor{darkblue}{rgb}{0,0.592,0.655}%{0, 151, 167}
\definecolor{darkyellow}{rgb}{0.655,0.608,0}%{167, 155, 0}
\usepackage{times,graphicx,tabulary,multirow,xspace,xcolor}

\usepackage{pifont}
\usepackage{changepage}

\newcommand\blfootnote[1]{%
  \begingroup
  \renewcommand\thefootnote{}\footnote{#1}%
  \addtocounter{footnote}{-1}%
  \endgroup
}
\newcommand{\modelname}{\textsc{MM-ReAct}}

\makeatletter
\DeclareRobustCommand\onedot{\futurelet\@let@token\@onedot}
\def\@onedot{\ifx\@let@token.\else.\null\fi\xspace}

\def\eg{\emph{e.g}\onedot} 
\def\ie{\emph{i.e}\onedot} 
 
\def\etc{\emph{etc}\onedot}

\usepackage[capitalize]{cleveref}
\crefname{section}{Sec.}{Secs.}
\Crefname{section}{Section}{Sections}
\Crefname{table}{Table}{Tables}
\crefname{table}{Tab.}{Tabs.}

\def\cvprPaperID{****} % *** Enter the CVPR Paper ID here
\def\confName{****}
\def\confYear{****}

\begin{document}

%%%%%%%%% TITLE
\title{\textbf{\modelname}~\includegraphics[height=15pt]{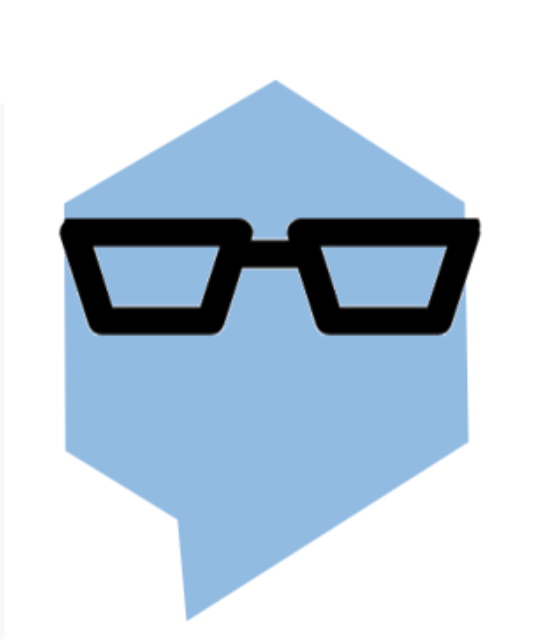}: Prompting ChatGPT for Multimodal Reasoning and Action}

% \author{ }
\author{Zhengyuan Yang$^{*}$, Linjie Li$^{*}$, Jianfeng Wang$^{*}$, Kevin Lin$^{*}$, Ehsan Azarnasab$^{*}$, Faisal Ahmed$^{*}$, \\ Zicheng Liu, Ce Liu, Michael Zeng, Lijuan Wang$^{\spadesuit}$ \\
Microsoft Azure AI \\ 
{\tt\footnotesize \{zhengyang,lindsey.li,jianfw,keli,ehazar,fiahmed,zliu,ce.liu,nzeng,lijuanw\}@microsoft.com}
}

\twocolumn[{
\renewcommand\twocolumn[1][]{#1}
\maketitle
\centering
% \small
\vspace{-2em}
\captionsetup{type=figure}
\includegraphics[width=1.\textwidth]{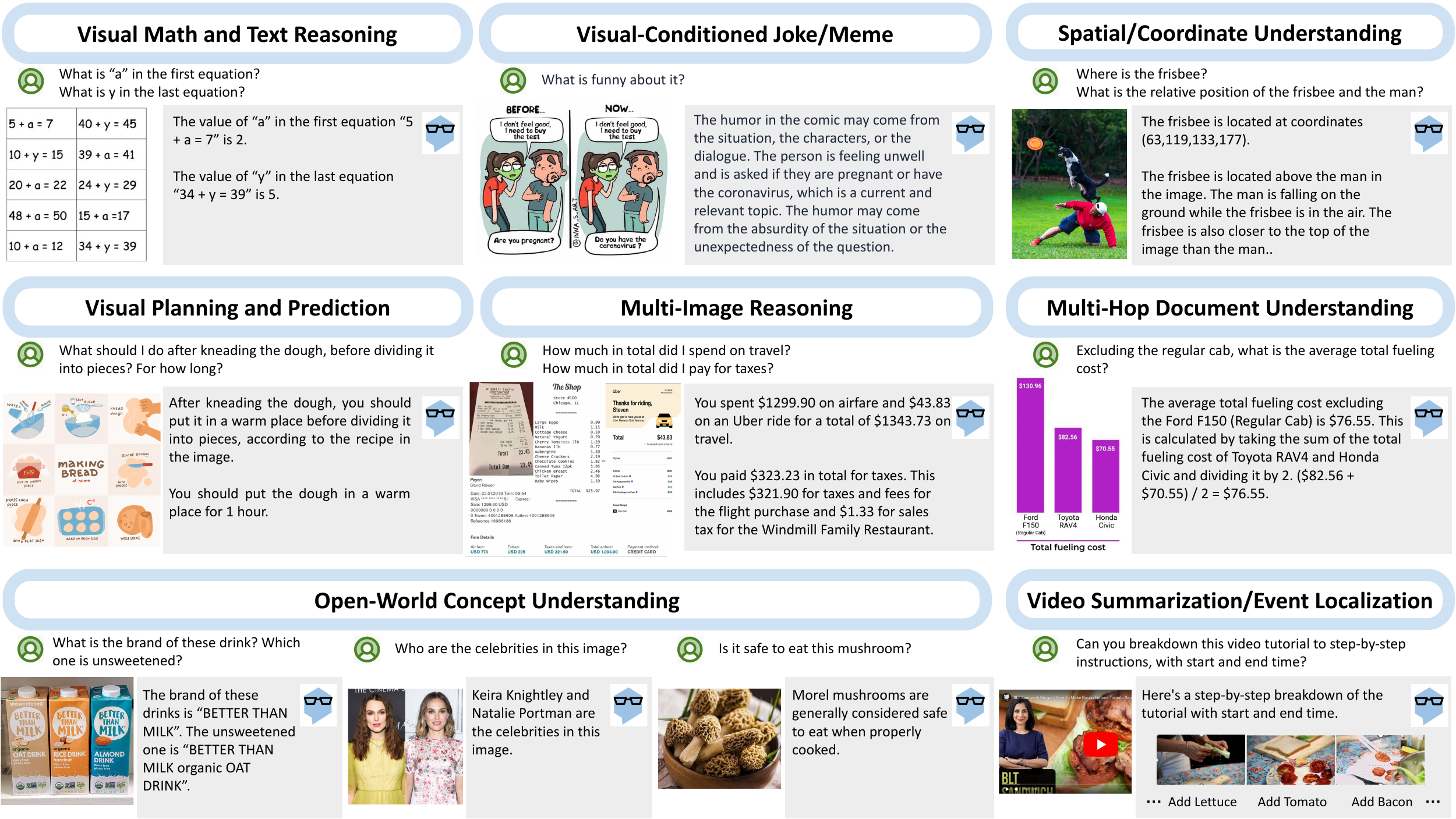}
\vspace{-1em}
\captionof{figure}{\modelname~allocates specialized vision experts with ChatGPT to solve challenging visual understanding tasks through multimodal reasoning and action. For example, the system could associate information from multiple uploaded receipts and calculate the total
travel cost (``Multi-Image Reasoning''). We only highlight key information here and postpone full \modelname~responses to Figures~\ref{fig:exp1_math}-\ref{fig:exp2_video}.
}
\label{fig:teaser}
\vspace{1.9em}
}
]

%%%%%%%%% BODY TEXT
\begin{abstract}
% \vspace{-2mm}
We propose \modelname, a system paradigm that integrates ChatGPT with a pool of vision experts to achieve multimodal reasoning and action. In this paper, we define and explore a comprehensive list of advanced vision tasks that are intriguing to solve, but may exceed the capabilities of existing vision and
vision-language models. To achieve such advanced visual intelligence, \modelname~introduces a textual prompt design that can represent text descriptions, textualized spatial coordinates, and aligned file names for dense visual signals such as images and videos. \modelname's prompt design allows language models to accept, associate, and process multimodal information, thereby facilitating the synergetic combination of ChatGPT and various vision experts. Zero-shot experiments demonstrate \modelname's effectiveness in addressing the specified capabilities of interests and its wide application in different scenarios that require advanced visual understanding. Furthermore, we discuss and compare \modelname's system paradigm with an alternative approach that extends language models for multimodal scenarios through joint finetuning.
Code, demo, video, and visualization are available at \url{https://multimodal-react.github.io/}.
% \vspace{0pt}
\blfootnote{$^*$Equal Contribution\hspace{3mm}$^\spadesuit$Project Lead}
\end{abstract}
\section{Introduction}
Recent years have seen significant advancement for computer vision, thanks to 
improved network architecture~\cite{he2016deep,vaswani2017attention,dosovitskiy2020image,carion2020end}, large-scale model training~\cite{yuan2021florence,dehghani2023scaling,wang2022git}, and other factors.
However, different vision problems typically require different models, 
which often require manual selection and composition of individual models for each use case.
For example, when determining if an image contains ``people'', we may choose the image tagging model~\cite{huiskes2008mir,chua2009nus,lin2014microsoft} and check if the predicted tag list contains \textit{``people''}. 
If \textit{``people''} exists, we may select the celebrity model~\cite{liu2015deep} to further understand 
whether a celebrity appears and who he/she is.

One research direction is to combine the vision and language modules as one end-to-end model,
such as Flamingo~\cite{alayrac2022flamingo}, PaLM-E~\cite{driess2023palme},
to provide a dialogue-based experience to the end user. 
That is, the user can use natural language to interact with the model around the image content.
The vision module encodes vision signals into special text tokens or features that the language module can understand, enabling the system to utilize the language module for understanding user queries and providing responses.
However, these joint finetuning approaches require a large amount of computing resources and annotated data to enable specific capabilities.
In this work, we aim to combine existing individual vision models with the language model in a more flexible manner to tackle complicated visual understanding problems, \eg, the ones illustrated in Figure~\ref{fig:teaser}. 

Large language models (LLMs)~\cite{brown2020language,chowdhery2022palm}, such as ChatGPT, have shown impressive dialogue capability with text as both input and output. Recent NLP research~\cite{yao2022react,gao2022pal,trivedi2022interleaving,schick2023toolformer} (\eg, \textsc{ReAct}~\cite{yao2022react}) demonstrates the effectiveness of integrating external NLP tools, such as search engines and math calculators, with LLMs by proper instruction. 
Specifically, \textsc{ReAct}~\cite{yao2022react} prompts an LLM to generate \textit{reasoning} texts that break down complex problems into intermediate steps, and \textit{action} texts that allocate NLP tools for solving these steps.
One example is that the LLM can suggest a text query to a modern search engine to grab the latest internet information, and return the user with the information that is not in the pre-training corpus. 
Inspired by the efficacy of reasoning and acting with LLMs and NLP tools, we explore the integration of vision expert tools with LLMs.

To this end, we present \modelname, a system paradigm that composes numerous vision experts with ChatGPT for multimodal reasoning and action.
To enable images and videos as inputs, we use their file path as the input to ChatGPT.
The file path functions as a placeholder, allowing ChatGPT to treat it as a black box.
Whenever a specific property such as celebrity names or box coordinates is required, ChatGPT is expected to seek help from a specific vision expert to identify the desired information. 
To inject the knowledge of vision experts' usages into ChatGPT, we add instructions to ChatGPT prompts about each expert's capability, input argument type, and output type, along with a few in-context examples for each expert. 
Additionally, a special watchword is instructed such that we can use regex expression matching to invoke the expert accordingly.

We show \modelname's representative visual understanding capabilities in Figure~\ref{fig:teaser}. For example, \modelname~could associate information from multiple uploaded receipts and calculate the total
travel cost (``Multi-Image Reasoning''), recognize and answer questions about the ``morel mushrooms'' (``Open-World Concept Understanding''), and condense a long video into representative thumbnails (``Video Summarization and Event Localization''). These visual intelligence features are similar to those found in recent models, such as multimodal GPT-4~\cite{gpt4} and PaLM-E~\cite{driess2023palme}, but achieved via prompting instead of additional multimodal training. The \modelname~system may provide extra flexibility in module upgrades, and may be effective in certain visual understanding tasks by better utilizing existing specialized vision experts, such as celebrity recognition and dense captioning.
\section{Related Work}
%%%%%%%%%%%%%%%%%%%%%%%%%%%%%%%%%%%%%%%
\noindent\textbf{LLMs Prompting Methods.}
Large language models (LLMs)~\cite{brown2020language,chowdhery2022palm} demonstrate a strong chain-of-thought (CoT) capability~\cite{wei2022chain,kojima2022large} that could break down complex problems into solvable intermediate steps. 
On the other hand, research~\cite{nakano2021webgpt,huang2022language,ahn2022can} shows that LLMs, when equipped with a range of external NLP tools, can effectively serve as action planners to select and utilize tools for problem-solving, such as using search or mathematical tools to address knowledge or math problems.

Nevertheless, LLMs for reasoning~\cite{wei2022chain,kojima2022large} and LLMs for action~\cite{nakano2021webgpt,huang2022language,ahn2022can} 
, when used independently,
fail to solve complex tasks that require breaking down the problem via reasoning and solving sub-steps via planned actions.
Recent studies~\cite{yao2022react,gao2022pal,trivedi2022interleaving,schick2023toolformer} have attempted to merge the action and reasoning phases to enhance LLMs' capabilities in solving complicated tasks that require advanced planning and reasoning. One representative work, \textsc{ReAct}~\cite{yao2022react}, treats reasoning text generation as an executable action and achieves the synergetic combination of reasoning and action for NLP tasks. In this work, we explore how to extend such intriguing properties into multimodal scenarios by modeling thought and invoking vision tools as executable actions.

\begin{figure*}[t]
\centering
\includegraphics[width=.99\textwidth]{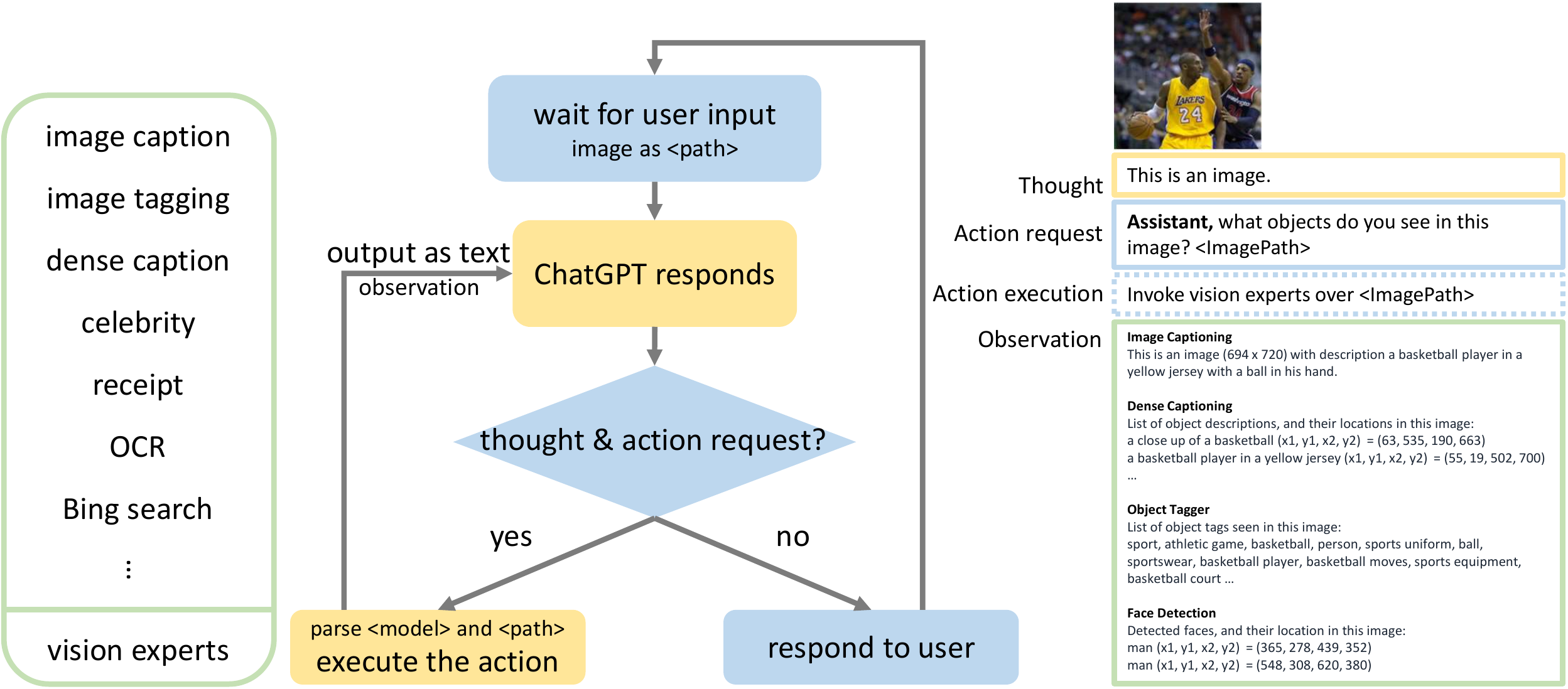}
\caption{Flowchart of \modelname~for enhanced visual understanding with ChatGPT. 
The user input can be in the form of text, images, or videos, with the latter two represented as file path strings.
ChatGPT is instructed to say specific watchwords in \textit{action request} if a vision expert is required to interpret the visual inputs. 
Regular expression matching is applied to parse the expert's name and the file path, which are then used to call the vision expert (\textit{action execution}).
The expert's output (\textit{observation}) is serialized as text and combined with the history to further activate ChatGPT.
If no extra experts are needed, \modelname~would return the final response to the user.
The right figure shows a single-round vision expert execution, which is the component that constructs the full execution flow illustrated in Figure~\ref{fig:prompt}.
}
\label{fig:arch}
\end{figure*}

\vspace{1mm}
\noindent\textbf{Vision+LLMs.}
Our \modelname~is related to the previous studies that extend language models to understand visual inputs. The representative framework adds a vision module to project visual inputs into representations that the language model can understand. These representations can be either discrete text words~\cite{yang2022empirical,zeng2022socratic,wanglanguage,hu2022promptcap} or continuous features projected into the textual feature space~\cite{tsimpoukelli2021multimodal,alayrac2022flamingo,li2023blip,huang2023language,driess2023palme}. 
Recent vision-language studies explore the chain-of-thought capability~\cite{wei2022chain,kojima2022large} in multimodal settings. MM-CoT~\cite{zhang2023multimodal} finetunes on the reasoning chain annotated in the ScienceQA~\cite{lu2022learn} dataset to achieve the CoT capability in the science question answering task. KOSMOS-1~\cite{huang2023language} and PaLM-E~\cite{driess2023palme} demonstrate the zero-shot multimodal CoT capabilities with large-scale training.

\vspace{1mm}
\noindent\textbf{Multimodal Reasoning and Action.} 
A key distinction between \modelname~and prior vision+LLM studies discussed above is that \modelname~leverages LLMs' high-level planning abilities to allocate various vision experts, rather than solely using LLMs for text generation conditioned on visual inputs.
\modelname~is closely related to the recent concurrent work of Visual ChatGPT~\cite{wu2023visual} and ViperGPT~\cite{suris2023vipergpt}. In comparison, Visual ChatGPT~\cite{wu2023visual} primarily focuses on image generation and editing, while our \modelname~mainly focuses on visual understanding. ViperGPT~\cite{suris2023vipergpt} instructs LLMs to generate Python code for a one-round query answering. In contrast, \modelname~is a multi-round, dialogue-based system that may integrate the strong QA model as one of its vision experts.
\section{\textbf{\modelname}~Prompting}

The goal of \modelname~is to compose numerous vision experts
to empower ChatGPT with visual understanding.
A vision expert is a computer vision model that takes an image as input and 
interprets the content from different perspectives.
For instance, the image captioning expert generates a natural description, the OCR expert extracts the scene text in the image, the celebrity recognition model identifies the celebrity names, and the object detection model extracts the salient object with bounding box locations.
At present, one may have to manually decide which vision experts to employ for specific use cases, and manually compose them. Instead, our target is to automate this process based on the requirements presented by the user query in natural language. 

ChatGPT is an artificial intelligence chatbot with text as both input and output, without visual understanding. However, ChatGPT exhibits strong instruct learning capability, which motivates us to instruct ChatGPT to properly determine which vision expert should be invoked and which image should be processed.

Figure~\ref{fig:arch} shows the flowchart of our \modelname~system. The terms \textit{thought} and \textit{action request} refer to the reasoning and action-oriented texts generated by ChatGPT to break down the problem or invoke vision experts. 
\textit{Observation} refers to the vision expert's responses after the \textit{action execution} requested in the \textit{action request} text.
% \textit{Observation} refers to the vision expert's responses after running the selected vision model (\ie, tools).
Next, we detail each step in the flowchart as follows.

\subsection{User Input}
As ChatGPT only accepts texts as input, the first challenge is how to accommodate non-text inputs, such as multiple images and videos.
Since most vision experts accept the file path or URL, we use the path string to indicate non-text inputs.
The file path itself is meaningless and is essentially a placeholder. 
Although no visual recognition task can be performed directly with file paths, ChatGPT may seek help from different vision experts to understand the image content from different perspectives, \eg, identifying the celebrity names of the detected person. 
By including the provided file path in its text output, ChatGPT can indicate which image should be processed by the vision expert. 

\subsection{ChatGPT Response}
Given the user's input, ChatGPT is expected to provide two kinds of responses. 
The first is to seek help from vision experts, while the second is to respond to the user directly. A key challenge is to set up a protocol such that we know when to invoke the vision expert. 
Inspired by \textsc{ReAct}~\cite{yao2022react}, we instruct ChatGPT to respond with certain watchwords, such as \textit{``Assistant, what objects are there in the image? $<$file path$>$''}, if a specific vision expert is required. In our implementation, we use the keyword  ``Assistant,'' to distinguish whether a vision expert is required.

To further improve the performance, 
we encourage ChatGPT to show the \textit{thought} (reasoning)
process to highlight why an external tool is required. 
It is also shown to be beneficial in NLP studies~\cite{yao2022react} to incorporate such reasoning.

\subsection{Vision Experts}
\label{sec:3.3}
Given the action request from ChatGPT, we use the regular expression matching to parse the expert name and the file path, and invoke the action (vision expert execution).

The expert's output can be in different forms but is standardized into the text format such that ChatGPT can understand it. For certain experts, such as the captioning model or the celebrity model, it is straightforward to represent the output as text. However, the standardization is less intuitive for others. For example, the detection model outputs a list of object names with bounding box locations. In this case, we concatenate all the boxes, each of which is represented as $<$object name, x1, y1, x2, y2$>$, where (x1,y1), (x2,y2) are the coordinates of the top-left and bottom-right corners, respectively. An additional text description is added to explain the meaning of the last four numerical values. In some cases, we find ChatGPT is capable of understanding these coordinates, \eg, identifying which object is on the left.

\begin{figure*}[t]
\centering
\vspace{-10mm}
\includegraphics[width=.9\textwidth]{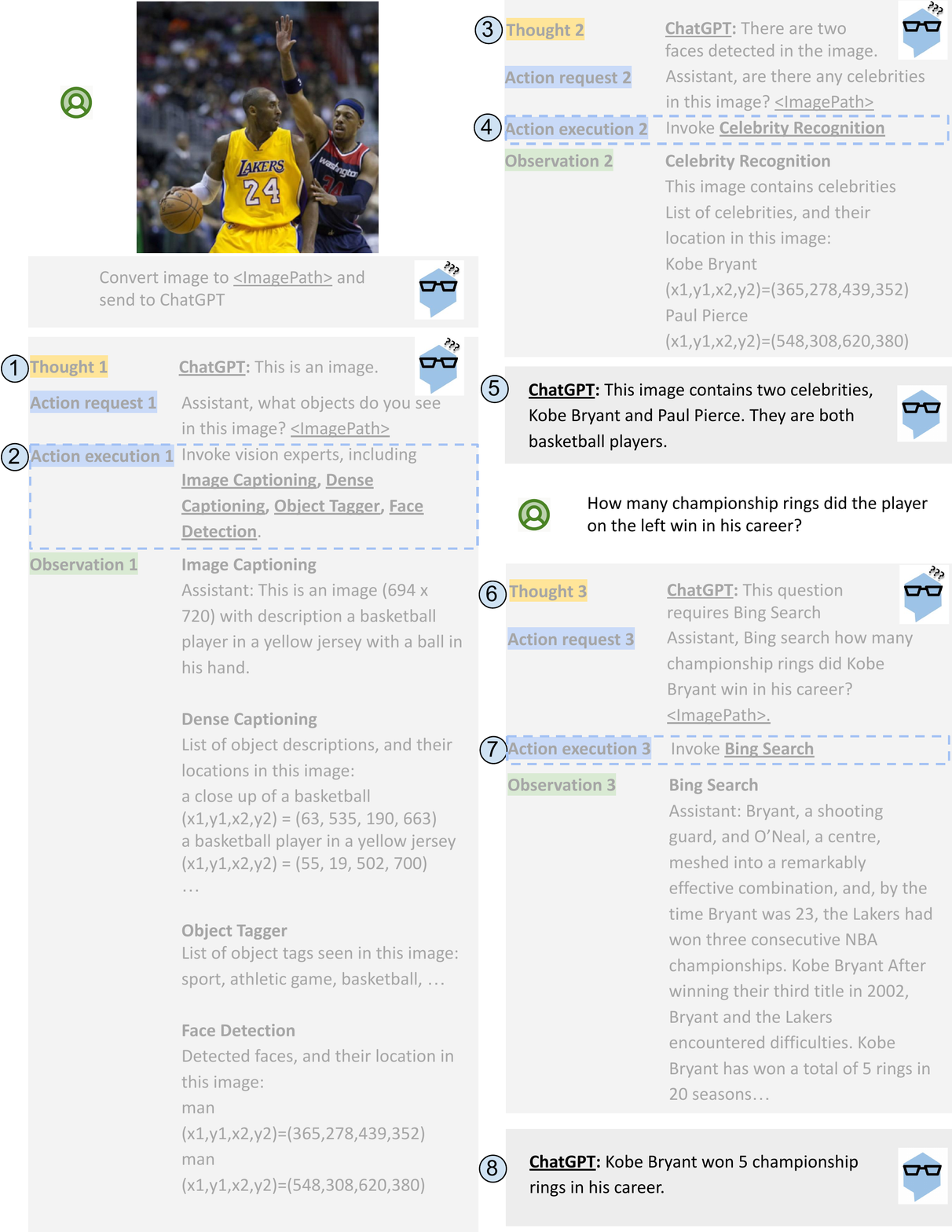}
\vspace{-1ex}
\caption{An example of \modelname's full execution flow. The blue circles with numbered indices indicate the order in which different models are called (\ie, the executions). The executions, highlighted by bold underlined text, can be either a ChatGPT call (\eg, ``\textbf{\underline{ChatGPT:}}'') or running one or multiple selected vision experts (\eg, ``\textbf{\underline{Image Captioning}}''). We add a commentary text \textit{action execution} in dashed boxes to help understand the expert execution. Each ChatGPT execution takes the preceding text as input and generates the text leading up to the next execution (\eg, \textit{``This is an image. Assistant, what $\ldots$ image? \textless ImagePath\textgreater''} for Execution~1).
Texts in gray represent \modelname's thoughts or vision experts' actions and outputs, which are invisible to users. This multimodal reasoning and action process occurs behind the scene to gather the necessary information for generating final responses to users, which are shown in black.
	}
\label{fig:prompt}
\end{figure*}

The text-formed output from vision experts can be interpreted as the \textit{observation} resulting from ChatGPT's action of invoking the vision expert.
Combining observations with the chat history, ChatGPT can further invoke additional experts or return the final answer to the user. We provide examples of full execution flows in Section~\ref{sec:4.2} and Figure~\ref{fig:prompt}.

To inject the knowledge of various vision experts' usages, we add both instructions and in-context examples in the prefix when prompting ChatGPT. 
Each expert is described with the model name, a general description of its capability, the input data format, and the output information. 
After describing each expert, we add a few in-context dialogue examples to enhance the performance.
With the injected knowledge, ChatGPT can effectively select one or multiple vision experts to understand the images or videos from different perspectives.

\subsection{Extensibility} 
Our scheme is motivated by \textsc{ReAct}, which invokes different tools in the NLP field. As only the text is involved, no specific design is required to incorporate other modalities. In this work, we extend \modelname~to the vision domain.
The key is to replace the non-text modality with a path string, enabling ChatGPT to ask specific vision experts to recognize the visual content.
Therefore, we could further extend \modelname~to other modalities, such as speech and audio.
Meanwhile, we can also easily incorporate more tools by formatting their outputs in a text format. 
While ChatGPT serves as the primary LLM in our main implementation, performance could be further enhanced through the simple upgrade to a more powerful LLM, such as GPT-4~\cite{gpt4} discussed in Section~\ref{sec:4.5}.
%%%%%%%%%%%%%%%%%%%%%%%%%%%%%%%%%%%%%%%%

%%%%%%%%%%%%%%%%%%%%%%%%%%%%%%%%%%%%%%%%
\section{Experiments}
\subsection{Experiment Setup}
%%%%%%%%%%%%%%%%%%%%%%%%%%%%%%%%%%%%%%%%
We implement \modelname~based on the LangChain codebase~\cite{langchain} and reference ideas from ReAct~\cite{yao2022react}. We access ChatGPT via the Azure ``gpt-3.5-turbo'' API that has a token length limit of 4,096, and utilize vision experts publicly available via the Azure Cognitive Services APIs\footnote{\url{https://azure.microsoft.com/en-us/products/cognitive-services/vision-services}}, including the ones for image captioning,
image tagging, dense captioning, optical character recognition (OCR), and specialized recognition models for celebrities, receipts, \etc. We further expand the toolset with customized tools for spatial understanding and image editing, and tools from other modalities such as Bing search and PAL math.

\subsection{\textbf{\modelname}'s Full Execution Flow}
\label{sec:4.2}
%%%%%%%%%%%%%%%%%%%%%%%%%%%%%%%%%%%%%%%%
Figure~\ref{fig:prompt} provides an example to illustrate \modelname's full execution flow.
We highlight the exact order to call different models (\ie, executions) with numbered blue circles. The executions, highlighted by bold underlined text, can be either a ChatGPT call (\eg, ``\textbf{\underline{ChatGPT:}}'') or the execution of one or multiple selected vision experts (\eg, ``\textbf{\underline{Image Captioning}}''). We add a commentary text \textit{action execution} in dashed boxes to help understand the vision expert execution. The \textit{action execution} is not an actual input or output in the \modelname~flow. ChatGPT executions can be used to generate thought (reasoning) and action texts that allocate vision experts~\includegraphics[height=12pt]{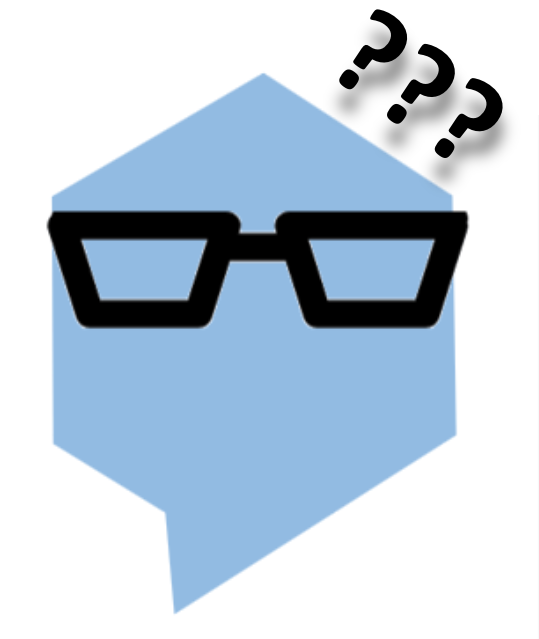} (invisible to users), or produce the final response to users~\includegraphics[height=12pt]{iccv2023AuthorKit/figure/placeholder/react.png}. Each ChatGPT execution takes the preceding text as input and generates the text leading up to the next execution (\eg, \textit{``This is an image. Assistant, what objects do you see
in this image? \textless ImagePath\textgreater''} for Execution~1). ChatGPT ``learns'' the proper text to generate based on the instructions and in-context examples in the prompt prefix, as detailed in Section~\ref{sec:3.3}.
Additional examples of the reasoning and execution procedures are in Figures~\ref{fig:mmreact-unfolded-2}-\ref{fig:mmreact-unfolded-4}.

\subsection{\textbf{\modelname}~Capabilities and Applications}
\begin{figure*}[t]
\centering
\includegraphics[height=.95\textheight]{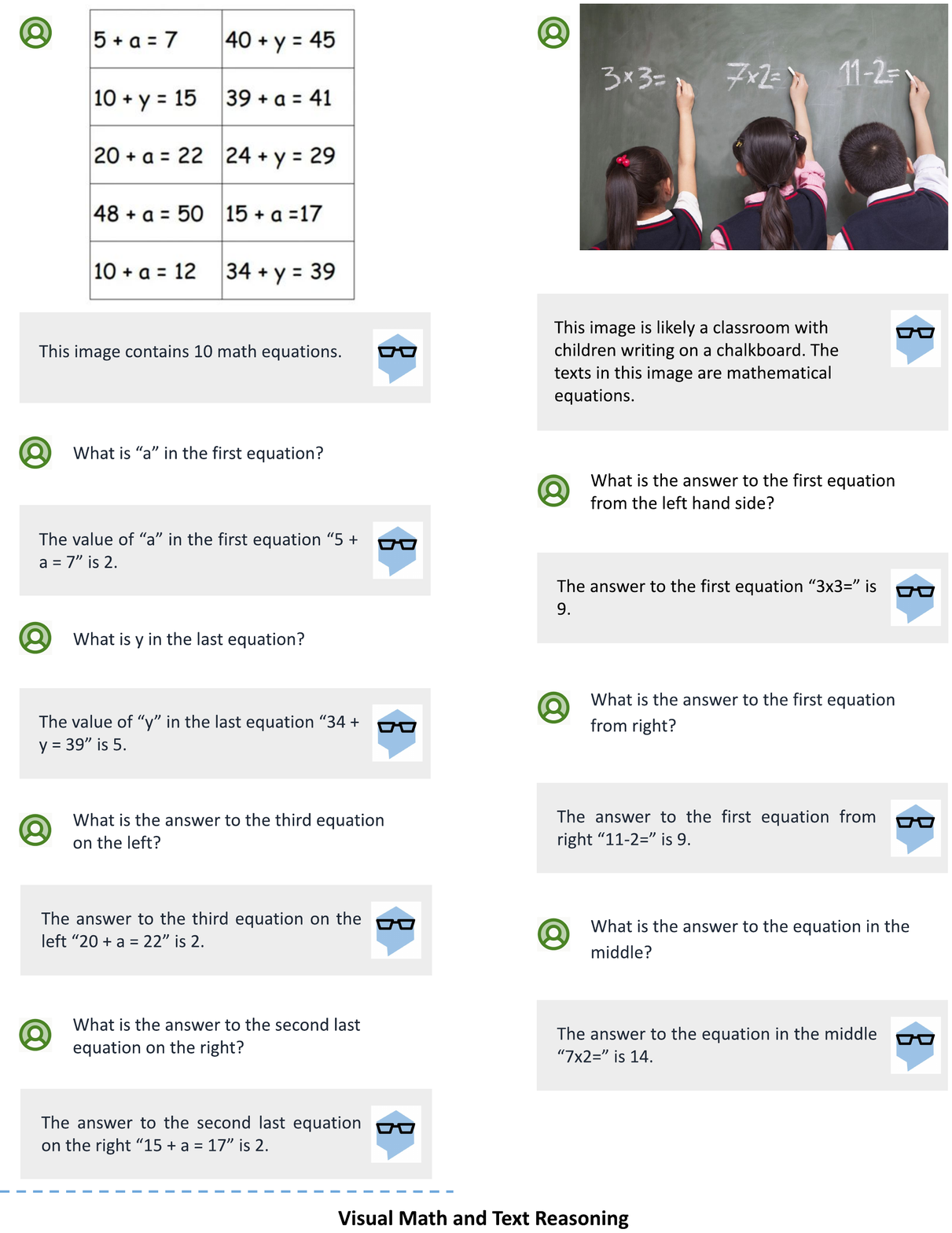}
\caption{Case studies of \modelname's capabilities and application scenarios: \textbf{visual math and text reasoning}. 
	}
\label{fig:exp1_math}
\end{figure*}

\begin{figure*}[t]
\centering
\includegraphics[height=.95\textheight]{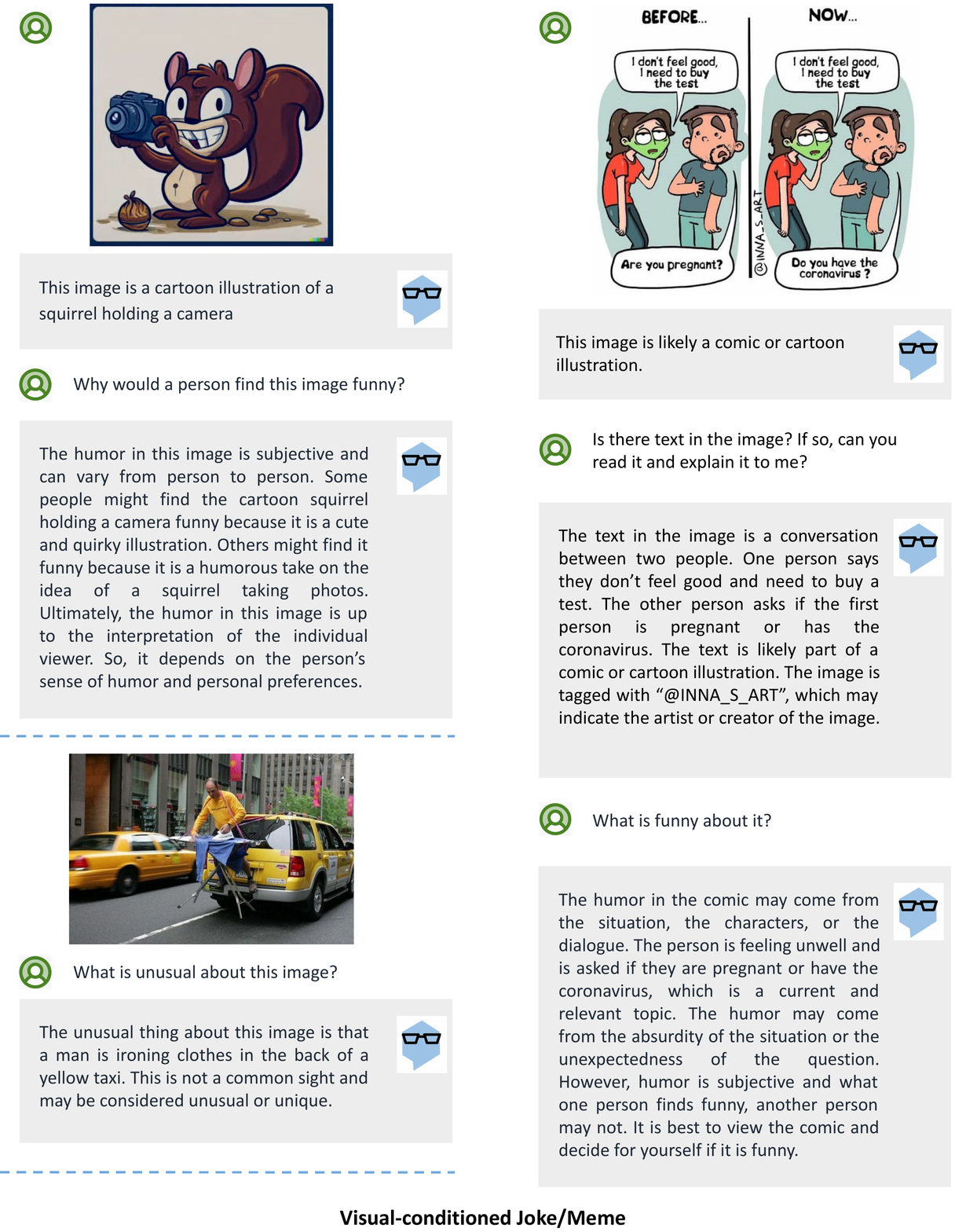}
\caption{Case studies of \modelname's capabilities and application scenarios:  \textbf{visual-conditioned joke/meme}. We provide an example of the unfolded multi-modal reasoning and action steps for meme understanding in Figure~\ref{fig:mmreact-unfolded-2}.
	}
\label{fig:exp1_meme}
\end{figure*}

\begin{figure*}[t]
\centering
\includegraphics[height=.95\textheight]{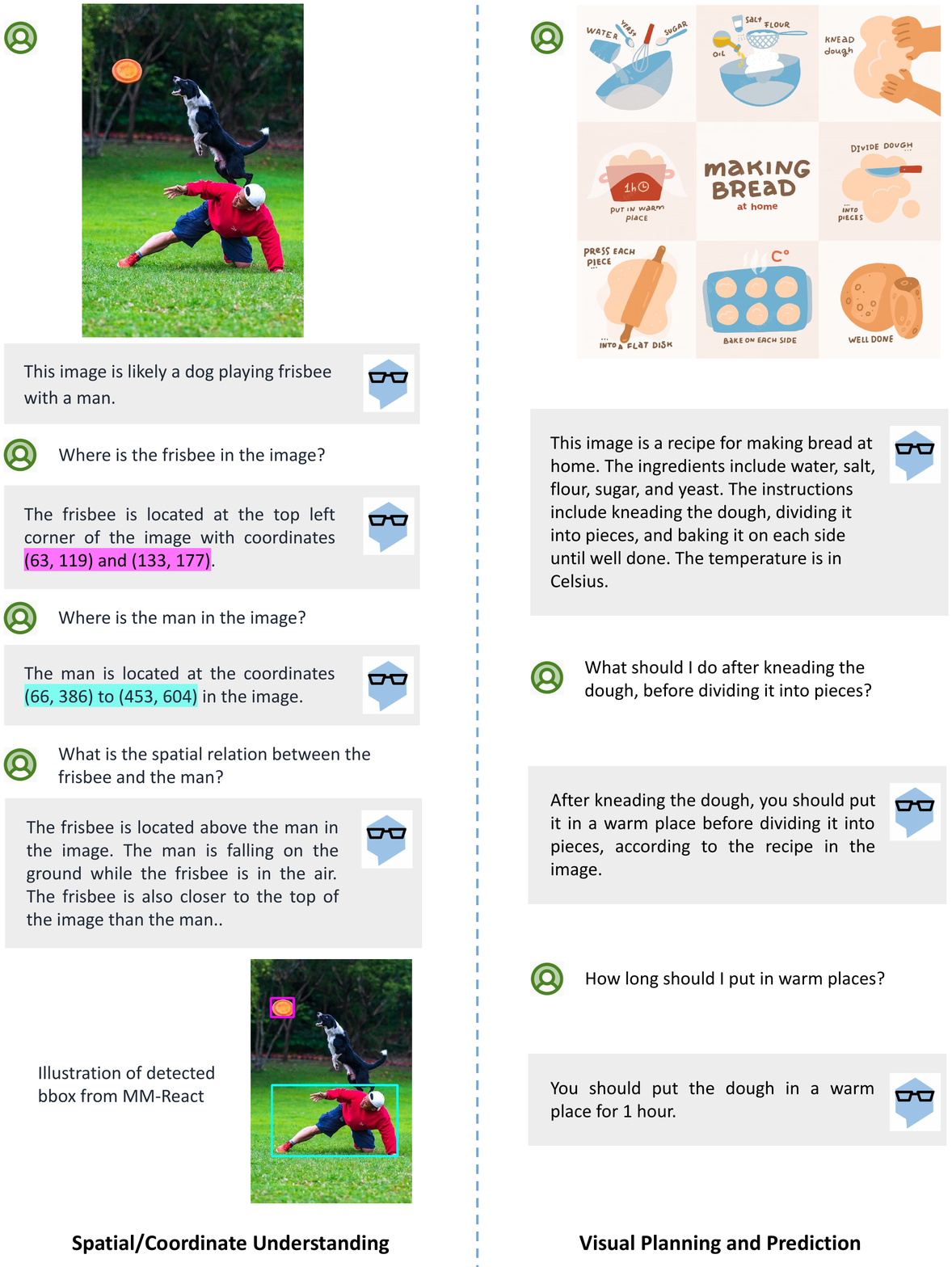}
\caption{Case studies of \modelname's capabilities and application scenarios: \textbf{spatial/coordinate understanding} and \textbf{visual planning and prediction}.
	}
\label{fig:exp1_spatial_plan}
\end{figure*}

\begin{figure*}[t]
\centering
\includegraphics[height=.95\textheight]{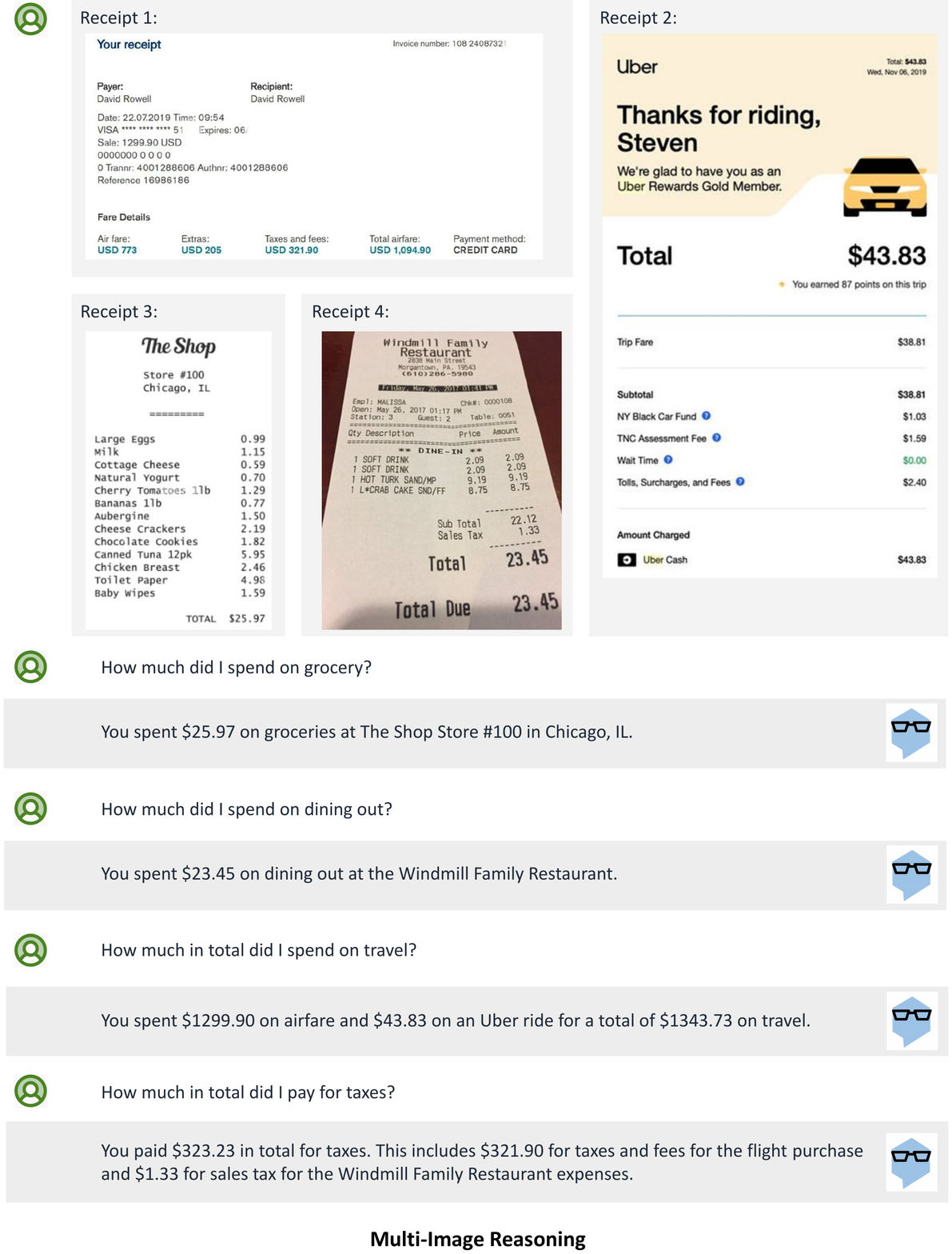}
\caption{Case studies of \modelname's capabilities and application scenarios: \textbf{multi-image reasoning}. For the multi-image inputs shown above, we input one receipt image at a time. Once all four receipt images are provided as inputs to  \modelname, we prompt it to answer questions that require reasoning over multiple images.
	}
\label{fig:exp1_multi_image3}
\end{figure*}

\begin{figure*}[t]
\centering
\includegraphics[height=.95\textheight]{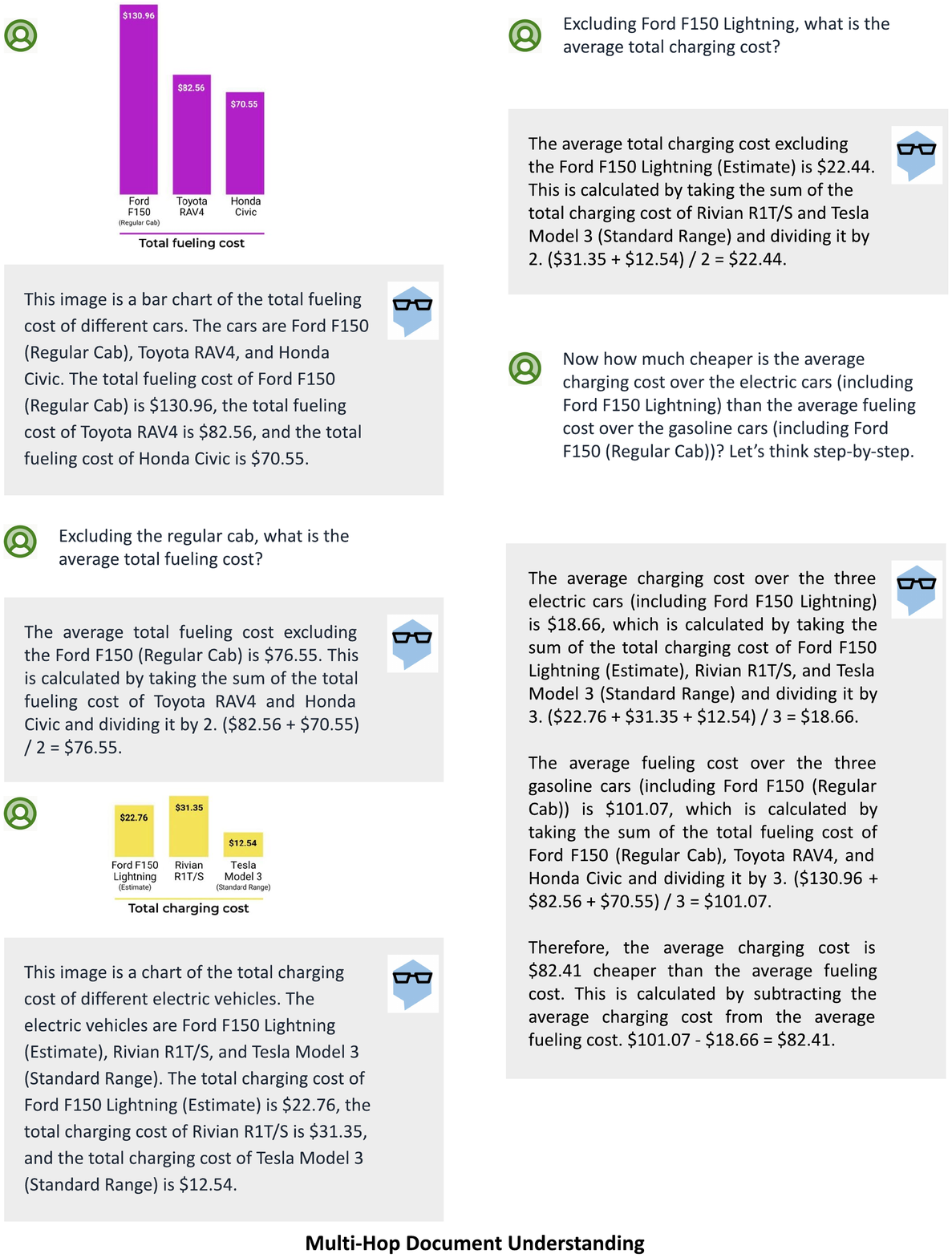}
\caption{Case studies of \modelname's capabilities and application scenarios: \textbf{multi-hop document understanding (bar charts)}. The unfolded multi-modal reasoning and action steps of this example are detailed in Figure~\ref{fig:mmreact-unfolded-chart}.
	}
\label{fig:exp1_multi_hop}
\end{figure*}

\begin{figure*}[t]
\centering
\includegraphics[height=.95\textheight]{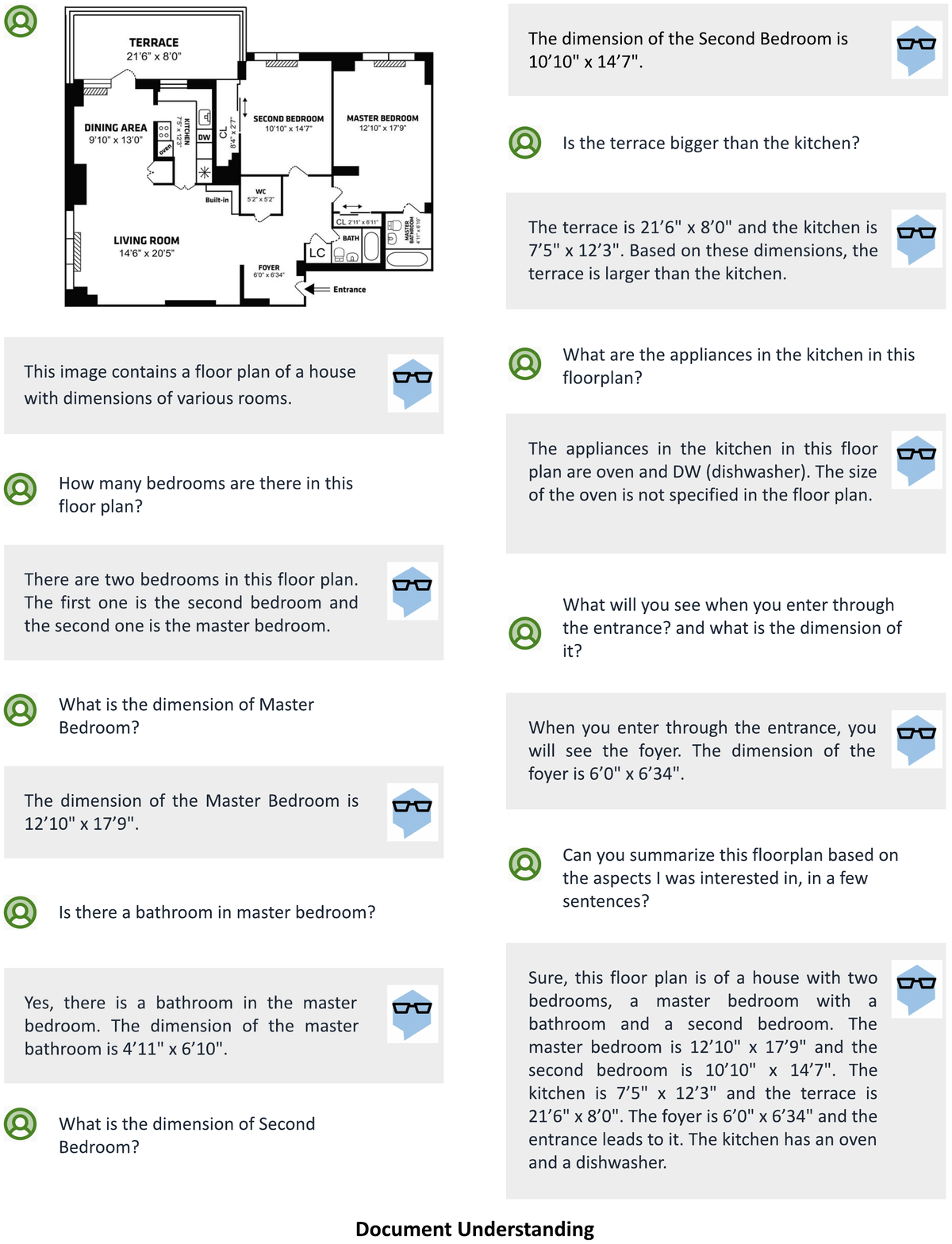}
\caption{Case studies of \modelname's capabilities and application scenarios: \textbf{document understanding (floorplan)}.
	}
\label{fig:exp1_floor}
\end{figure*}

\begin{figure*}[t]
\centering
\includegraphics[height=.95\textheight]{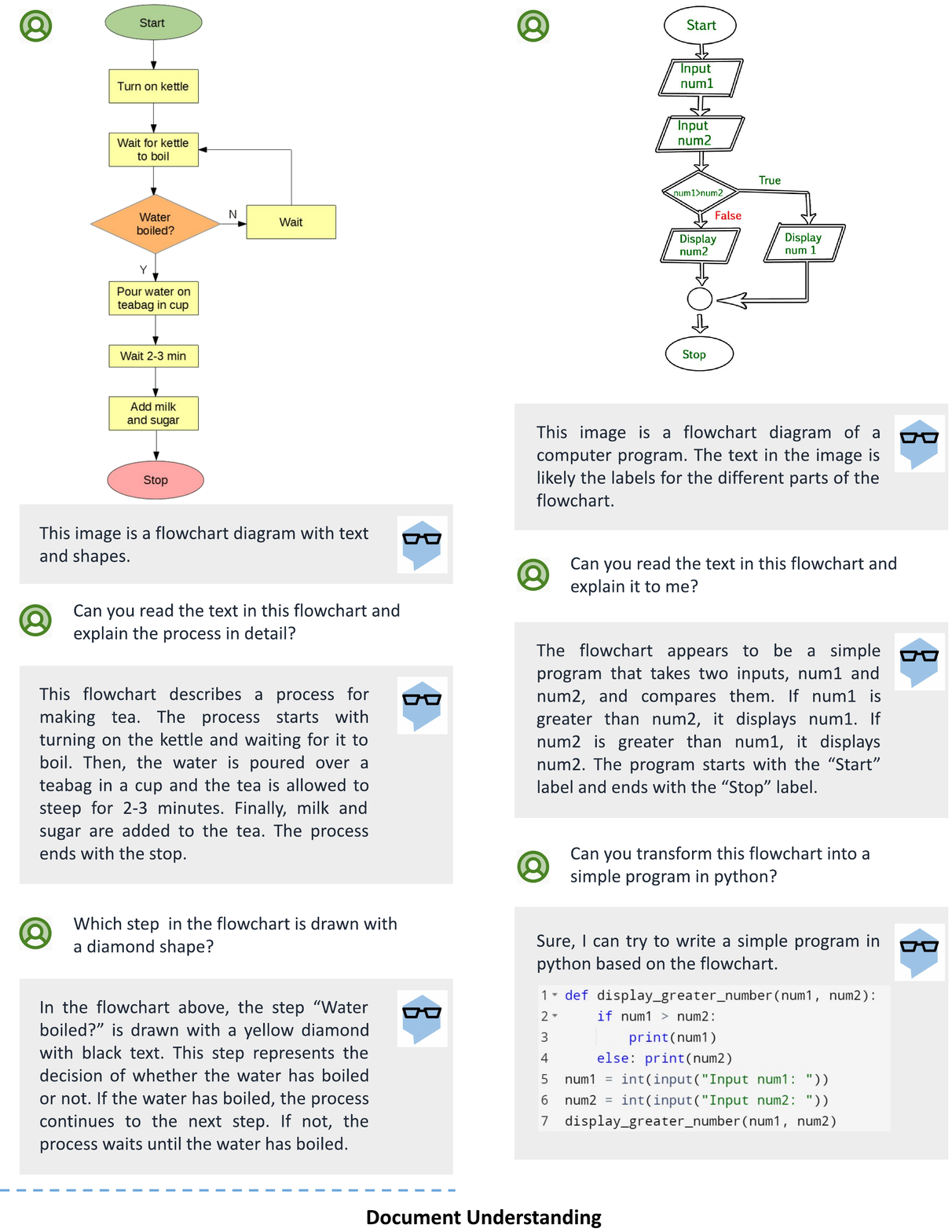}
\caption{Case studies of \modelname's capabilities and application scenarios: \textbf{document understanding (flowchart)}. We provide an example of the unfolded multi-modal reasoning and action steps for flowchart understanding in Figure~\ref{fig:mmreact-unfolded-flowchart}.
	}
\label{fig:exp1_flow}
\end{figure*}

\begin{figure*}[t]
\centering
\includegraphics[height=.95\textheight]{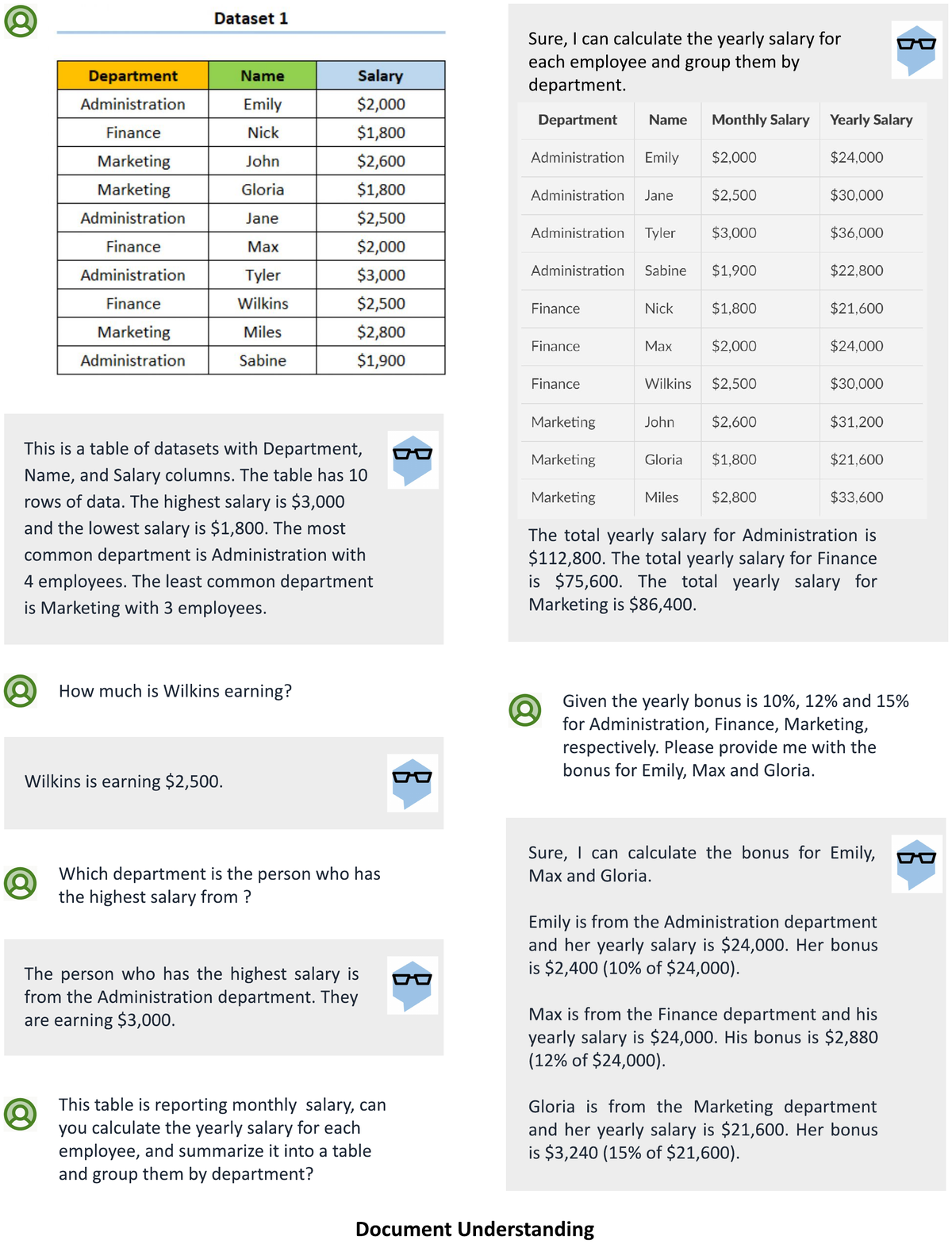}
\caption{Case studies of \modelname's capabilities and application scenarios: \textbf{document understanding (table)}.
	}
\label{fig:exp1_table}
\end{figure*}

\begin{figure*}[t]
\centering
\includegraphics[height=.95\textheight]{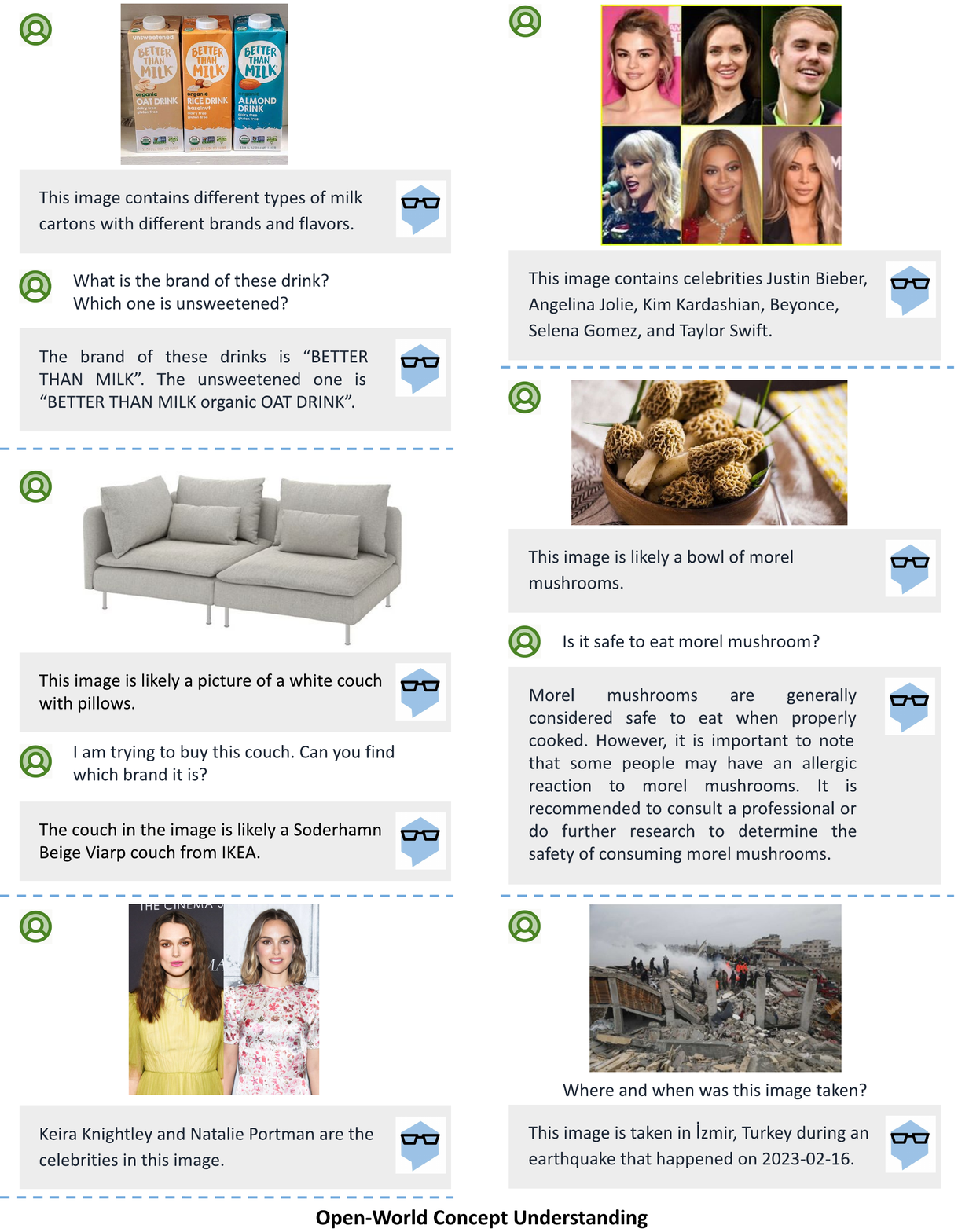}
\caption{Case studies of \modelname's capabilities and application scenarios: \textbf{open-world concept understanding}.
	}
\label{fig:exp1_open_world}
\end{figure*}

\begin{figure*}[t]
\centering
\vspace{-20mm}
\includegraphics[height=1.08\textheight]{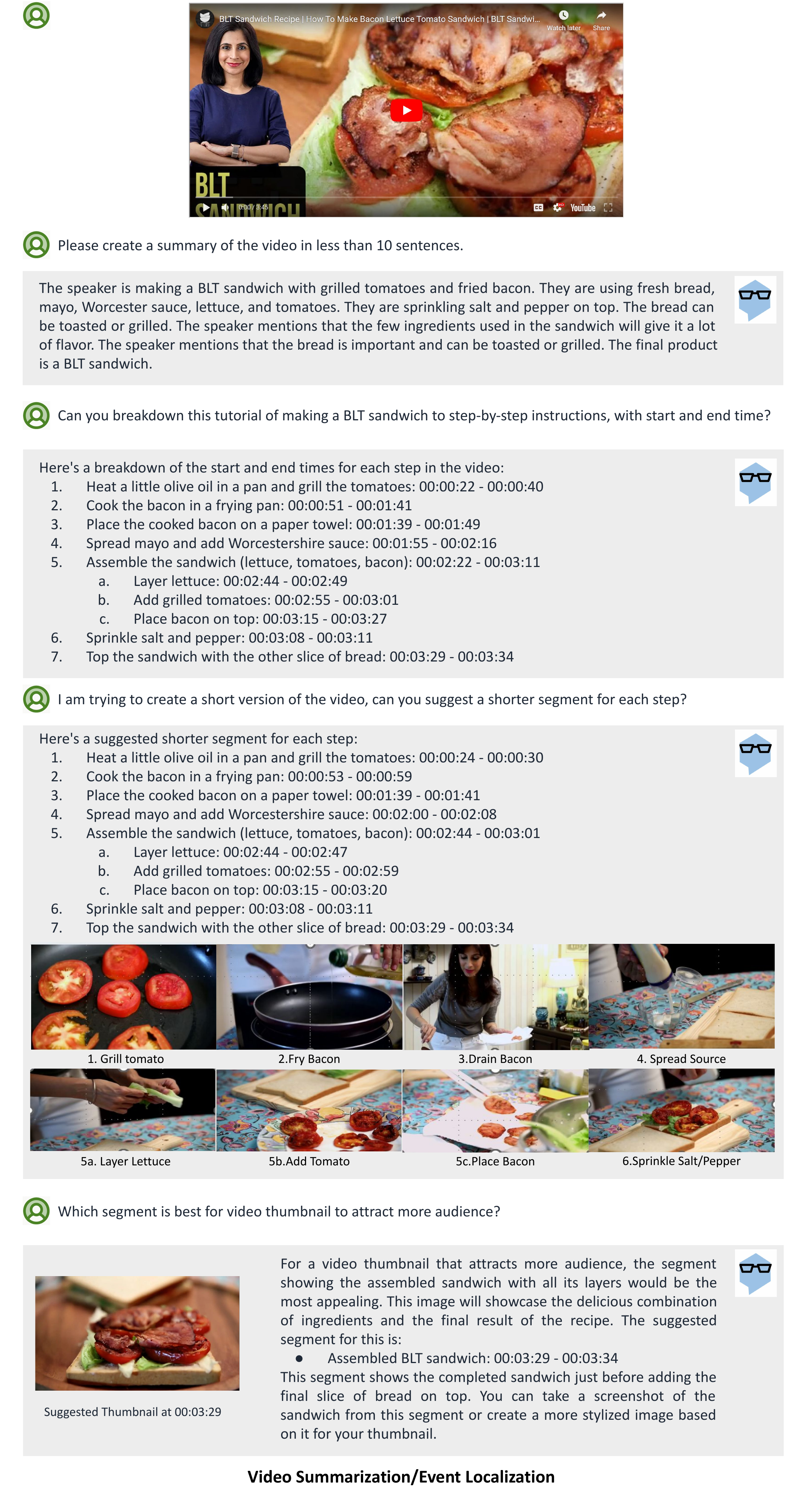}
\vspace{-5mm}
\caption{Case studies of \modelname's capabilities and application scenarios: \textbf{video summarization/event localization}.
	}

\label{fig:exp1_video}
\end{figure*}

\begin{figure*}[t]
\centering
\includegraphics[height=.95\textheight]{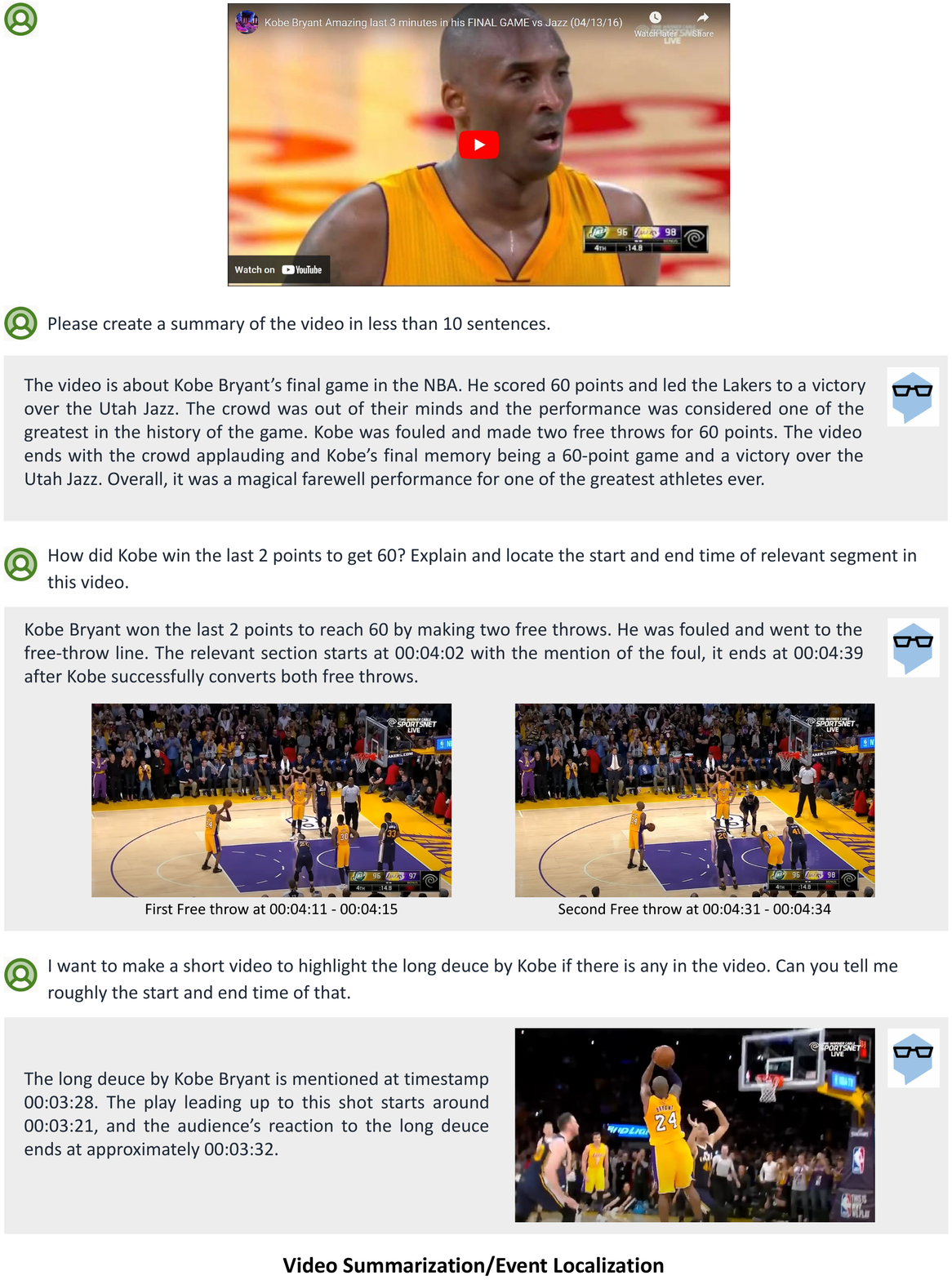}
\caption{Case studies of \modelname's capabilities and application scenarios: \textbf{video summarization/event localization}. 
	}
\label{fig:exp2_video}
\end{figure*}

\begin{figure*}[t]
\centering
\vspace{-20mm}
\includegraphics[width=1.\textwidth]{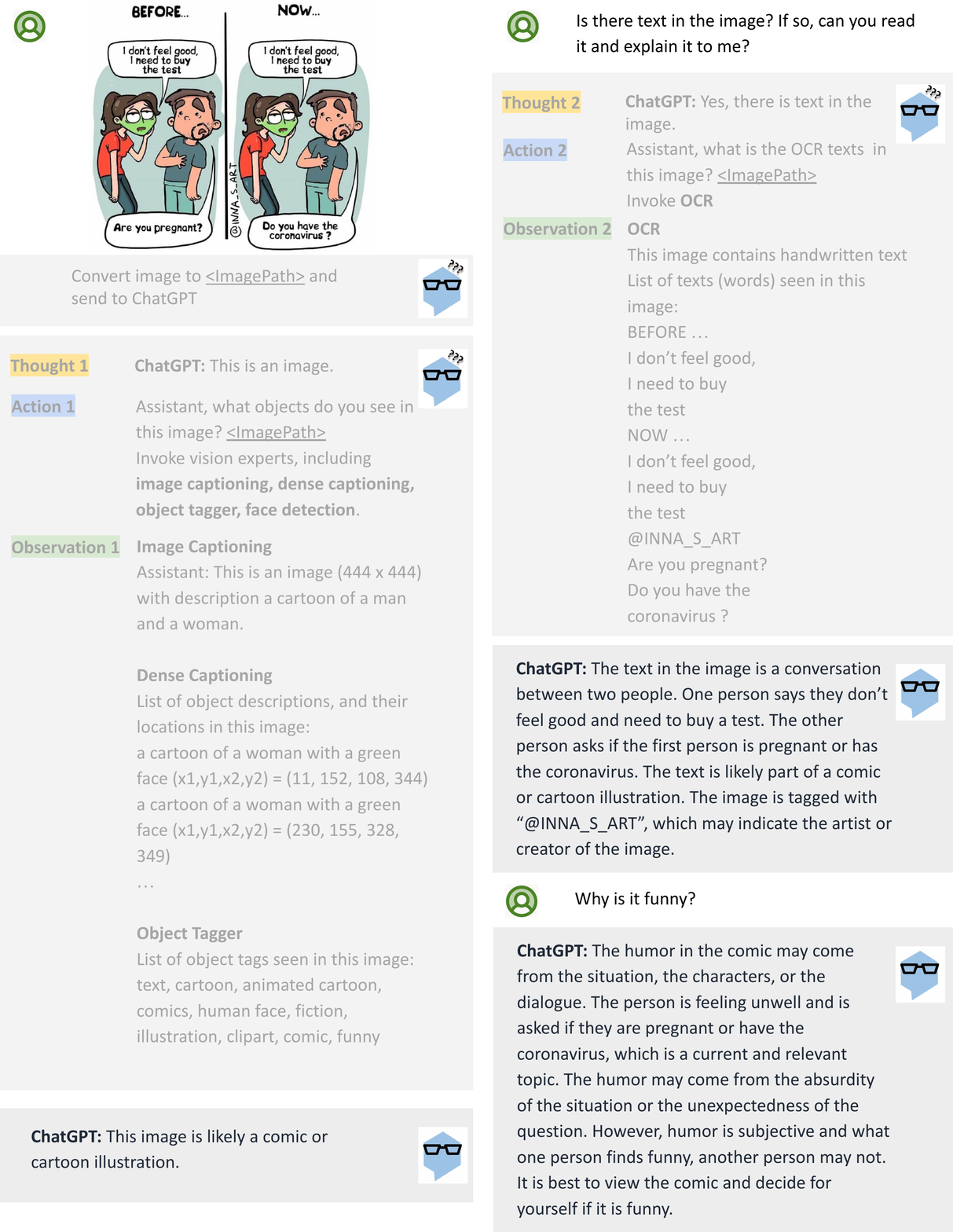}
\vspace{-0mm}
\caption{Unfolded multimodal reasoning and action steps for example of visual-conditioned joke/meme in Figure~\ref{fig:exp1_meme}.
	}
\label{fig:mmreact-unfolded-2}
\end{figure*}

\begin{figure*}[t]
\centering
\vspace{-20mm}
 \begin{adjustwidth}{-1cm}{-1cm}
 \centering
\includegraphics[width=1.15\textwidth]{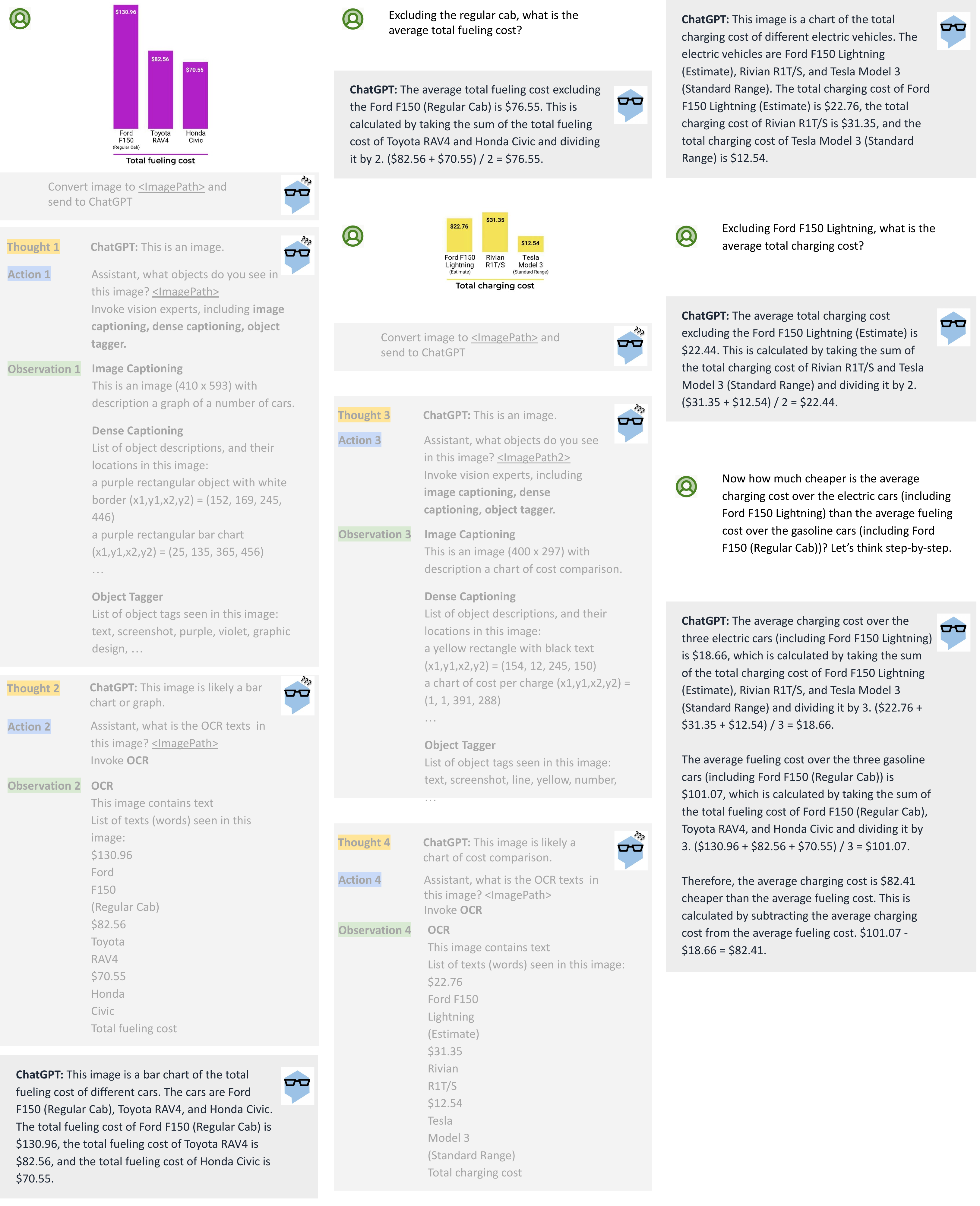}
\end{adjustwidth}
\vspace{-5mm}
\caption{Unfolded multimodal reasoning and action steps for multi-hop document understanding (bar charts) in Figure~\ref{fig:exp1_multi_hop}. 
	}
\label{fig:mmreact-unfolded-chart}
\end{figure*}

\begin{figure*}[t]
\centering
\vspace{-15mm}
\includegraphics[width=1\textwidth]{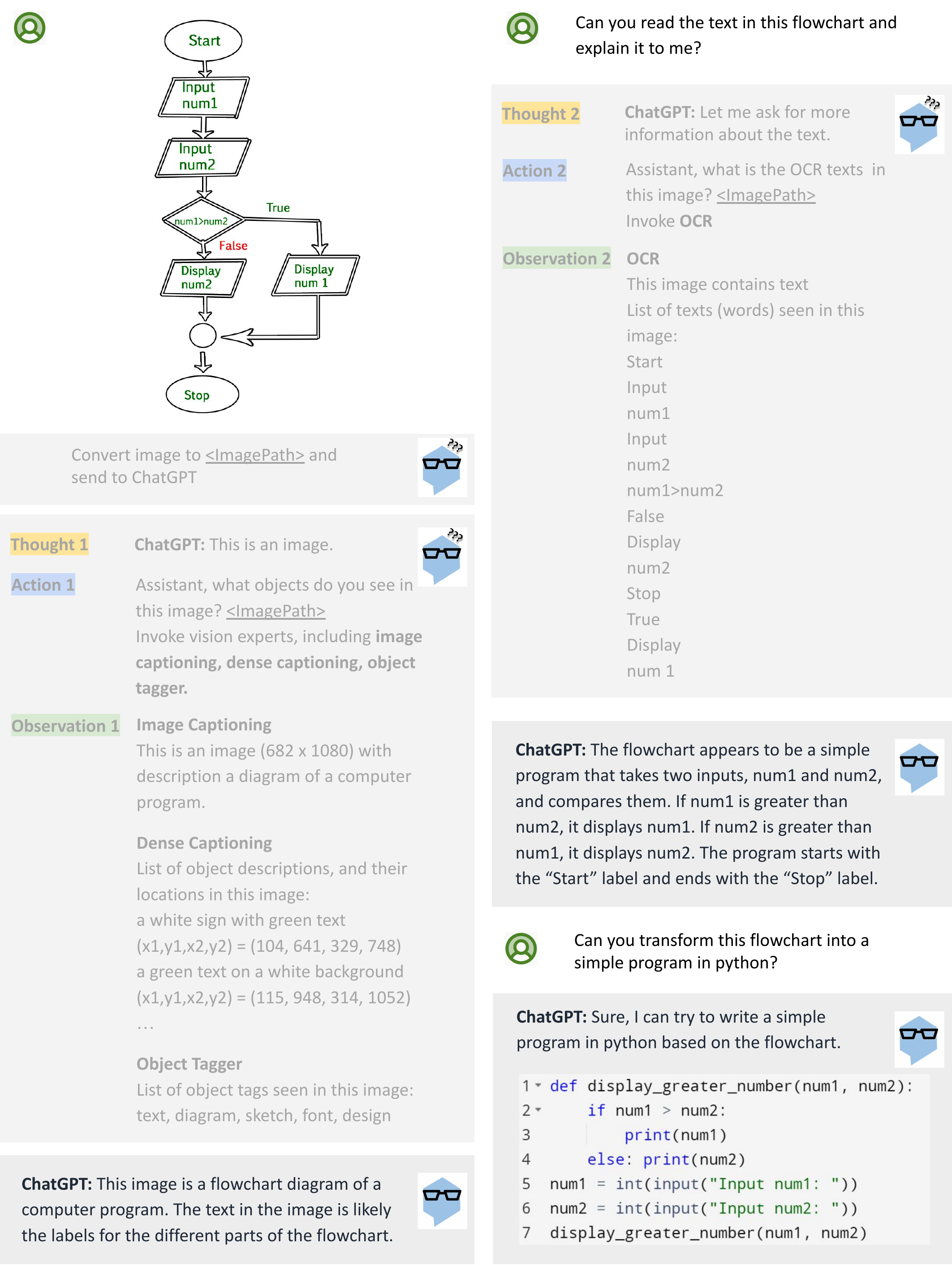}
\vspace{-5mm}
\caption{Unfolded multimodal reasoning and action steps for document understanding (flowchart) in Figure~\ref{fig:exp1_flow}.
	}
\label{fig:mmreact-unfolded-flowchart}
\end{figure*}

%%%%%%%%%%%%%%%%%%%%%%%%%%%%%%%%%%%%%%%%
Figures~\ref{fig:exp1_math}-\ref{fig:exp2_video} show the representative capabilities and application scenarios that \modelname~demonstrates. Specifically, we examine \modelname's capabilities in visual math and text reasoning (Figure~\ref{fig:exp1_math}), understanding visual-conditioned jokes and memes (Figure~\ref{fig:exp1_meme}), spatial and coordinate understanding, visual planning and prediction (Figure~\ref{fig:exp1_spatial_plan}), multi-image reasoning (Figure~\ref{fig:exp1_multi_image3}),  multi-hop document understanding on bar charts (Figure~\ref{fig:exp1_multi_hop}), floorplans (Figure~\ref{fig:exp1_floor}), flowcharts (Figure~\ref{fig:exp1_flow}),  tables (Figure~\ref{fig:exp1_table}), open-world concept understanding (Figure~\ref{fig:exp1_open_world}), and video analysis and summarization (Figure~\ref{fig:exp1_video},~\ref{fig:exp2_video}). We provide an example of the unfolded steps in Figure~\ref{fig:mmreact-unfolded-2}.

%%%%%%%%%%%%%%%%%%%%%%%%%%%%%%%%%%%%%%%%
\subsection{Capability Comparison with PaLM-E}
\begin{figure*}[t]
\centering
\vspace{-20mm}
\includegraphics[width=1.\textwidth]{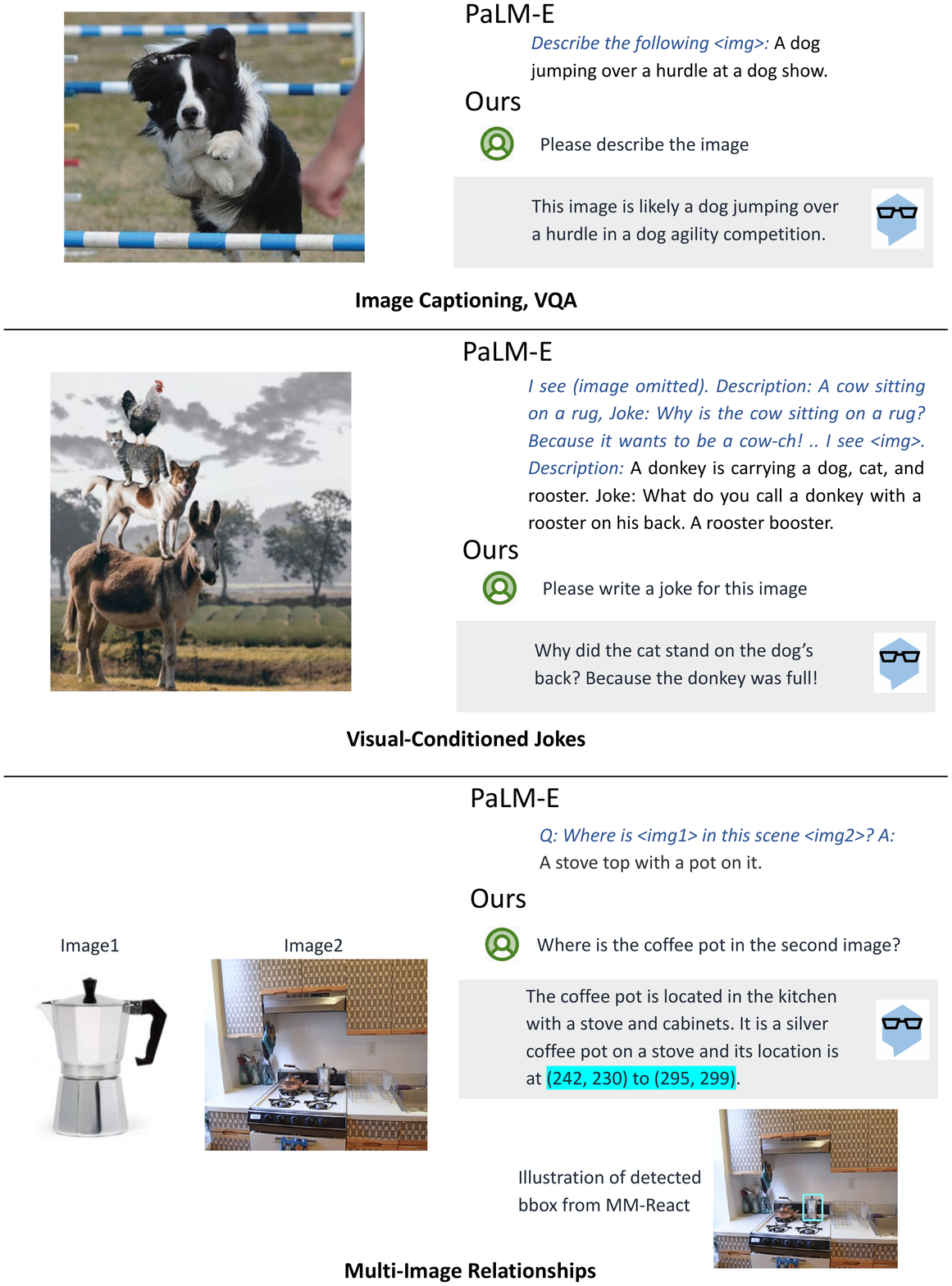}
\caption{Comparison of \modelname~with PaLM-E~\cite{driess2023palme} on illustrated capabilities. We empirically show that text prompts are as effective as expensive joint fine-tuning in solving complicated vision problems. 
	}
\label{fig:palme1}
\end{figure*}

\begin{figure*}[t]
\centering
\vspace{-20mm}
\includegraphics[width=1.\textwidth]{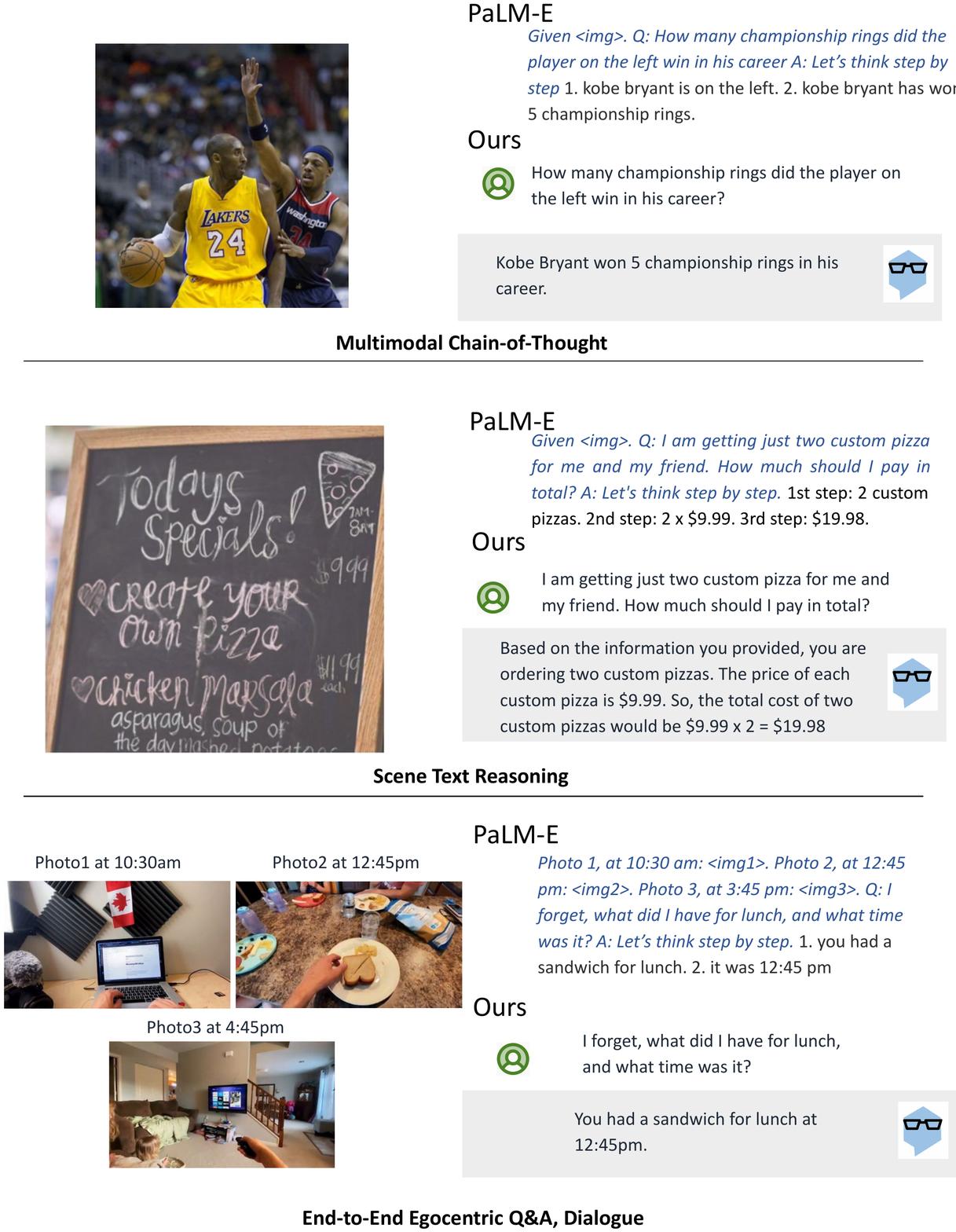}
\caption{Comparison of \modelname~with PaLM-E~\cite{driess2023palme} on illustrated capabilities. We empirically show that text prompts are as effective as expensive joint fine-tuning in solving complicated vision problems. 
	}
\label{fig:palme3}
\end{figure*}

\begin{figure*}[t]
\centering
\vspace{-20mm}
\includegraphics[width=1.\textwidth]{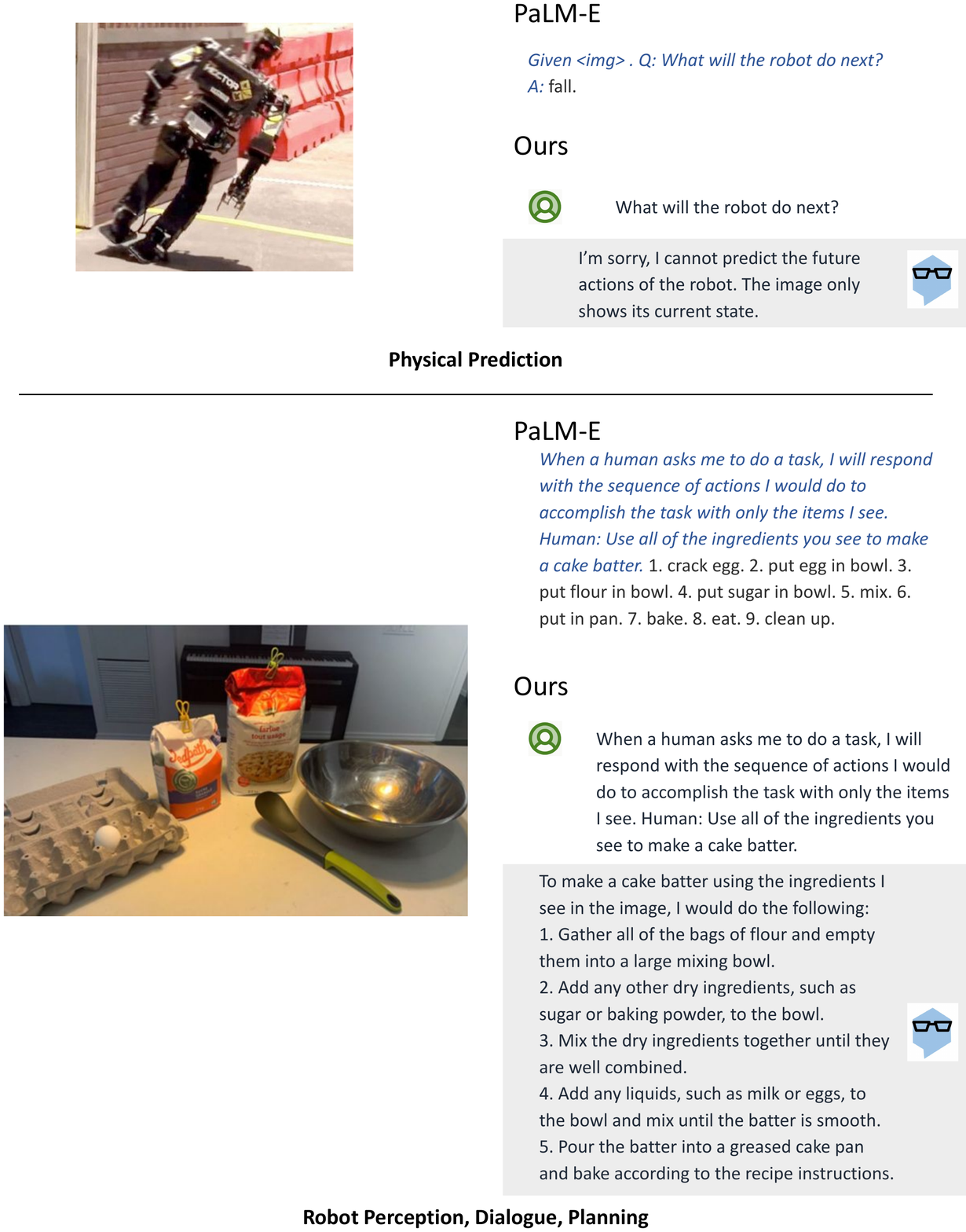}
\caption{Comparison of \modelname~with PaLM-E~\cite{driess2023palme} on illustrated capabilities. We empirically show that text prompts are as effective as expensive joint fine-tuning in solving complicated vision problems.
	}
\label{fig:palme4}
\end{figure*}

\begin{figure*}[t]
\centering
\vspace{-20mm}
\includegraphics[width=1.\textwidth]{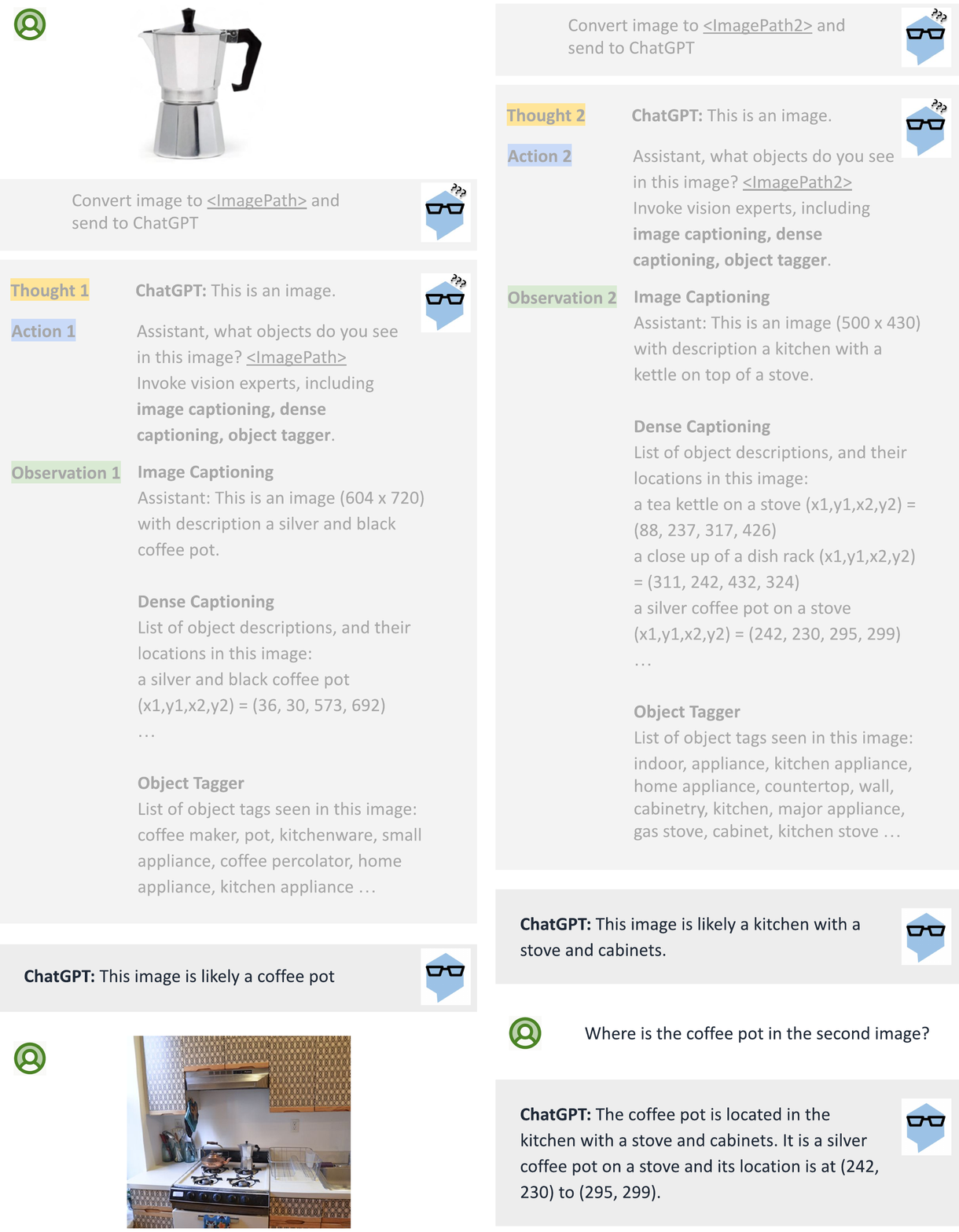}
\vspace{-0mm}
\caption{Unfolded multimodal reasoning and action steps for example of multi-image relationships in Figure\ref{fig:palme1}.
	}
\label{fig:mmreact-unfolded-3}
\end{figure*}

\begin{figure*}[t]
\centering
\vspace{-10mm}
\includegraphics[width=1.\textwidth]{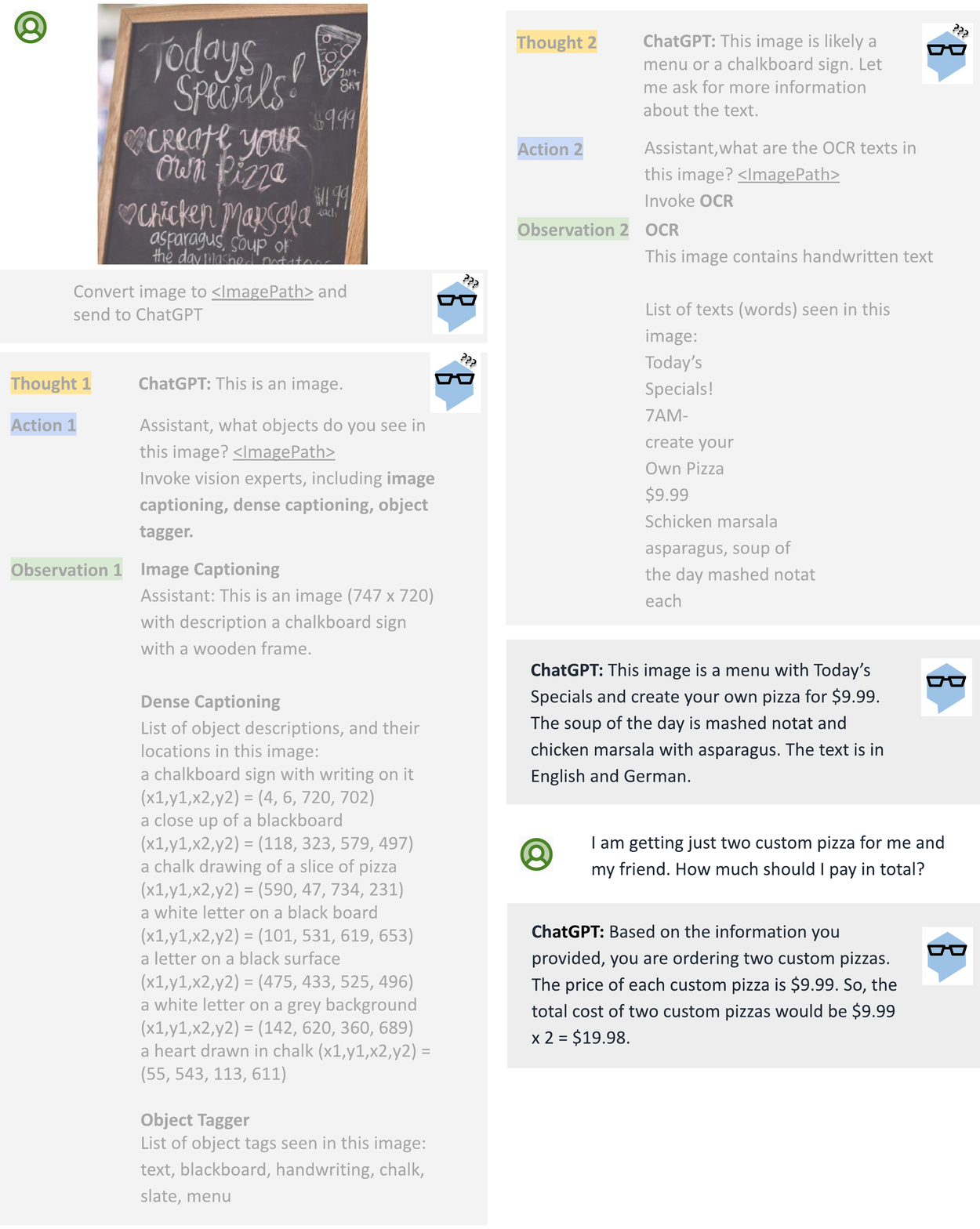}
\vspace{-0mm}
\caption{Unfolded multimodal reasoning and action steps for example of scene text reasoning in Figure~\ref{fig:palme3}.
	}
\label{fig:mmreact-unfolded-4}
\end{figure*}

\modelname~is a training-free scheme which composes existing vision experts with ChatGPT,
while PaLM-E~\cite{driess2023palme} trains a vision-language model which combines an image encoder and a text decoder with dedicated datasets.
Figures~\ref{fig:palme1}-\ref{fig:palme4} shows our \modelname~can achieve
competitive results to PaLM-E.
We further illustrate the complete multimodal reasoning and action procedures in Figures~\ref{fig:mmreact-unfolded-3},\ref{fig:mmreact-unfolded-4}.

\subsection{\textbf{\modelname}~Extensibility}
\label{sec:4.5}
\begin{figure*}[t]
\centering
\vspace{-25mm}
\begin{adjustwidth}{-1cm}{-1cm}
\centering
\includegraphics[height=1.1\textheight]{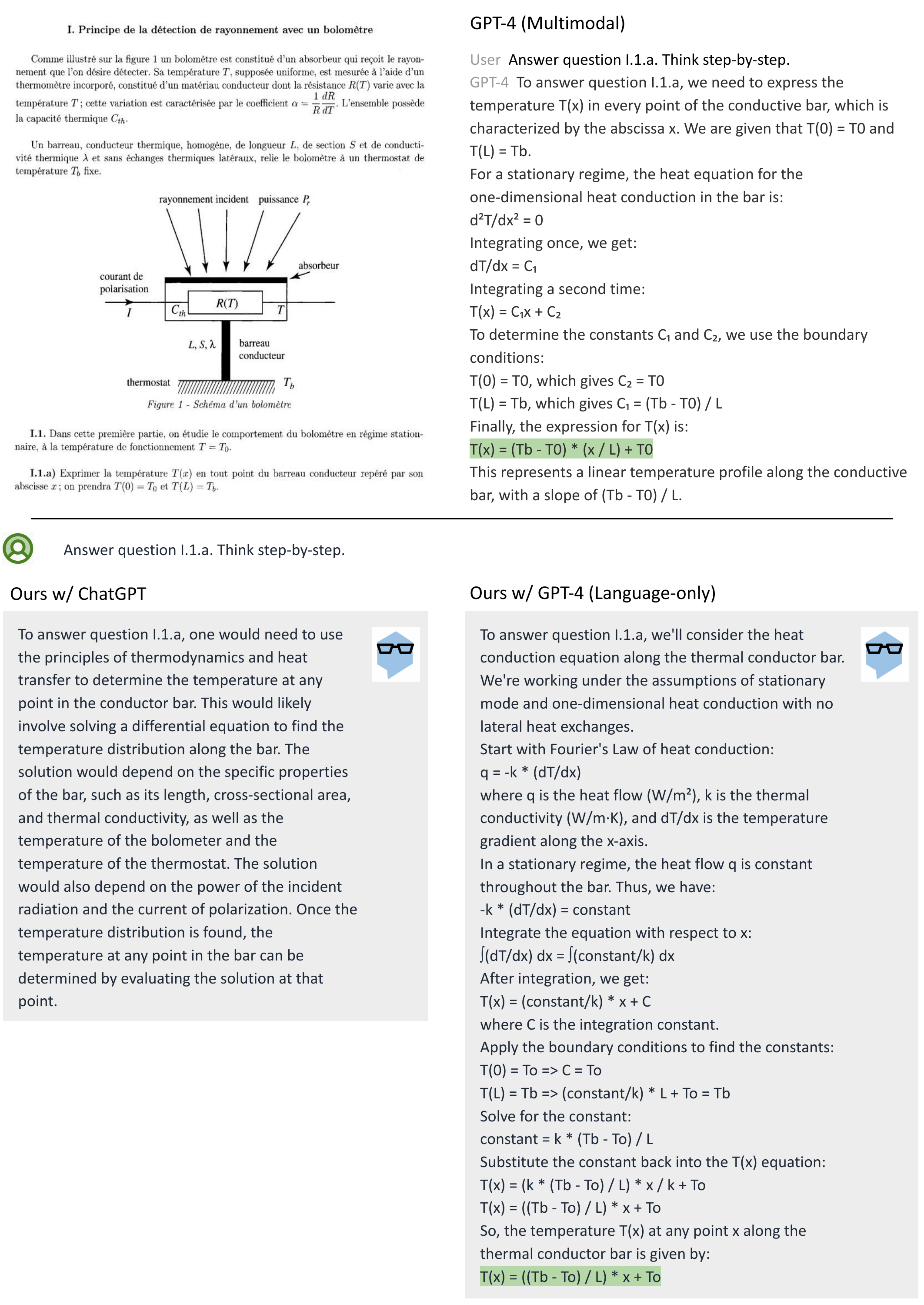}
\end{adjustwidth}
\vspace{-5mm}
\caption{Case studies of \modelname's \textbf{extensibility}. With the new release of GPT-4 (Language-only), we can upgrade ChatGPT in \modelname~to GPT-4 (Langauge-only) and further improve the results of \modelname. In Figure~\ref{fig:mmreact-unfolded-physics}, we provide the unfolded multimodal reasoning and action steps based on Ours w/ ChatGPT.
	}
\label{fig:chatgpt_gpt4_1}
\end{figure*}

\begin{figure*}[t]
\centering
\includegraphics[width=\textwidth]{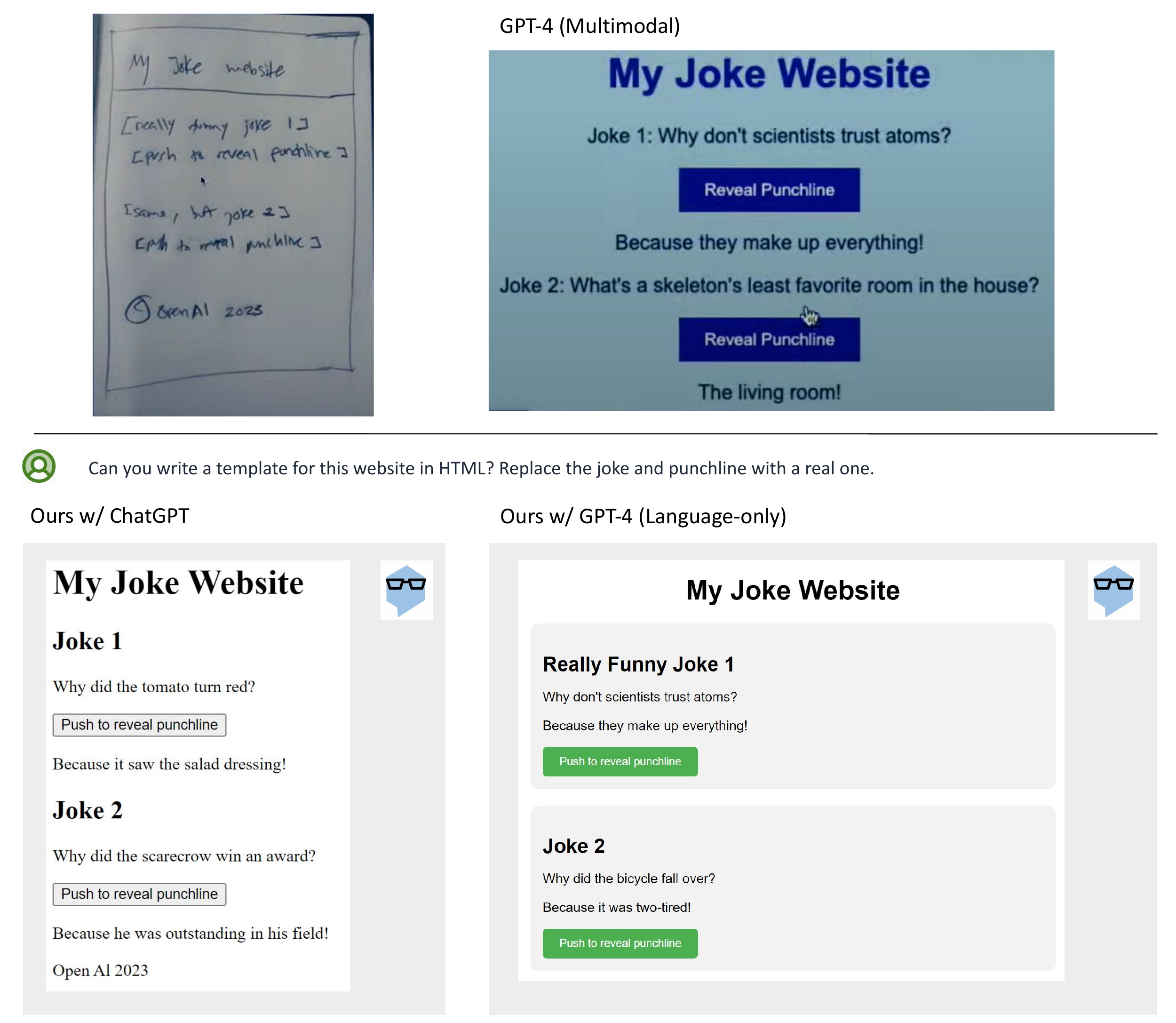}
\caption{Case studies of \modelname's \textbf{extensibility}. With the new release of GPT-4 (Language-only), we can upgrade ChatGPT in \modelname~to GPT-4 (Langauge-only)  and further improve the results of \modelname.
	}
\label{fig:chatgpt_gpt4_2}
\end{figure*}

\begin{figure*}[t]
\centering
\includegraphics[width=.98\textwidth]{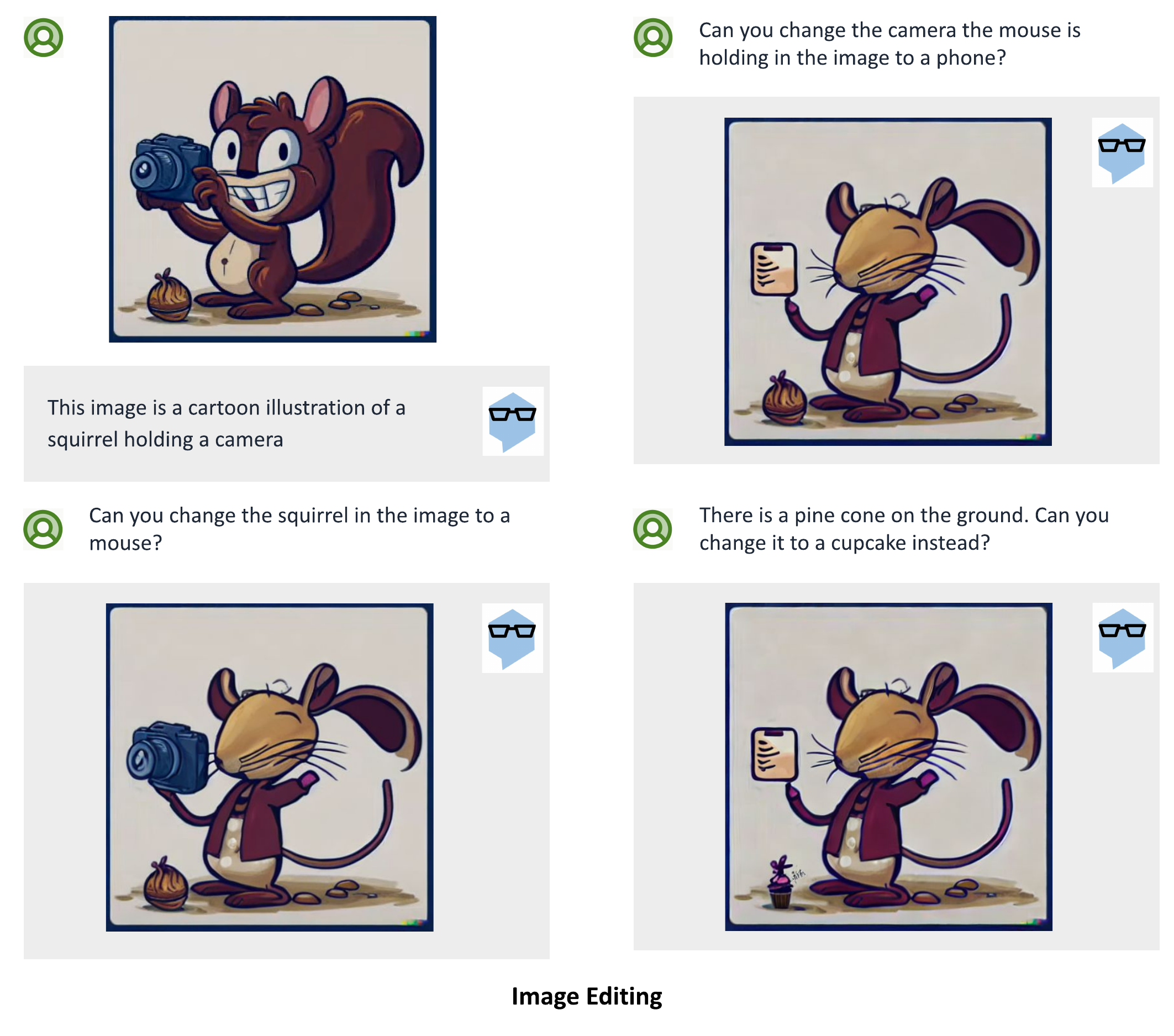}
\caption{Case studies of \modelname's \textbf{extensibility}. \modelname~is designed with the flexibility to plug in new experts. In this example, \modelname~is enhanced with the image editing model from X-decoder~\cite{zou2022xdecoder} and can perform dialogue-based image editing.
	}
\label{fig:ext_image_edit}
\end{figure*}

In Figure~\ref{fig:chatgpt_gpt4_1} and~\ref{fig:chatgpt_gpt4_2}, we explore the enhancement of \modelname's LLM from ChatGPT (``gpt-3.5-turbo'') to GPT-4 (language-only). We access the language-only GPT-4 via the ChatGPT website and reference the multimodal GPT-4 demo provided by OpenAI for comparison. These examples demonstrate the benefit of \modelname's extensibility: \modelname~equipped with GPT-4 correctly answers the physics question (Figure~\ref{fig:chatgpt_gpt4_1}), while the version with ChatGPT (GPT-3.5) fails. Furthermore, \modelname~is designed with the flexibility to incorporate new tools without training. Figure~\ref{fig:ext_image_edit} provides a case study of plugging an image editing tool from X-decoder~\cite{zou2022xdecoder} for multi-round, dialogue-based image editing.

%%%%%%%%%%%%%%%%%%%%%%%%%%%%%%%%%%%%%%%%
\subsection{Limitations}
We identify the following limitations. 
1). Considering the recognition capability in the wild, we find it hard to systematically evaluate the performance  with concrete accuracy numbers, due to 
a lack of annotated benchmarks. 
Thus, it is worth investing efforts towards how to effectively evaluate such system's performance.
2). The vision capability is limited by the integrated vision
experts. On one hand, the integrated experts may make mistakes; on the other hand, the system may fail if the necessary experts are missing. 
3). We inject the vision experts' knowledge in the prefix, and thus the number of experts is limited by the context window (4096 tokens) of ChatGPT. 
4). Visual signals are converted to text words
for ChatGPT understanding, which  might not be the optimal solution for certain vision tasks. 
5). \modelname~requires manual prompt engineering. We expect research work to automate this process, making the system even easier to develop.


%%%%%%%%%%%%%%%%%%%%%%%%%%%%%%%%%%%%%%%%
%%%%%%%%%%%%%%%%%%%%%%%%%%%%%%%%%%%%%%%%
%%%%%%%%%%%%%%%%%%%%%%%%%%%%%%%%%%%%%%%%
%%%%%%%%%%%%%%%%%%%%%%%%%%%%%%%%%%%%%%%%
%%%%%%%%%%%%%%%%%%%%%%%%%%%%%%%%%%%%%%%%
%%%%%%%%%%%%%%%%%%%%%%%%%%%%%%%%%%%%%%%%
%%%%%%%%%%%%%%%%%%%%%%%%%%%%%%%%%%%%%%%%
%%%%%%%%%%%%%%%%%%%%%%%%%%%%%%%%%%%%%%%%
%%%%%%%%%%%%%%%%%%%%%%%%%%%%%%%%%%%%%%%%
%%%%%%%%%%%%%%%%%%%%%%%%%%%%%%%%%%%%%%%%
%%%%%%%%%%%%%%%%%%%%%%%%%%%%%%%%%%%%%%%%
%%%%%%%%%%%%%%%%%%%%%%%%%%%%%%%%%%%%%%%%
%%%%%%%%%%%%%%%%%%%%%%%%%%%%%%%%%%%%%%%%
%%%%%%%%%%%%%%%%%%%%%%%%%%%%%%%%%%%%%%%%
%%%%%%%%%%%%%%%%%%%%%%%%%%%%%%%%%%%%%%%%
%%%%%%%%%%%%%%%%%%%%%%%%%%%%%%%%%%%%%%%%

%%%%%%%%%%%%%%%%%%%%%%%%%%%%%%%%%%%%%%%%
%%%%%%%%%%%%%%%%%%%%%%%%%%%%%%%%%%%%%%%%

%%%%%%%%%%%%%%%%%%%%%%%%%%%%%%%%%%%%%%%%
%%%%%%%%%%%%%%%%%%%%%%%%%%%%%%%%%%%%%%%%

%%%%%%%%%%%%%%%%%%%%%%%%%%%%%%%%%%%%%%%%
%%%%%%%%%%%%%%%%%%%%%%%%%%%%%%%%%%%%%%%%

%%%%%%%%%%%%%%%%%%%%%%%%%%%%%%%%%%%%%%%%
\begin{figure*}[t]
\centering
 \begin{adjustwidth}{-1cm}{-1cm} % Adjust the margins by -1cm on each side
 \centering
 \includegraphics[width=1.15\textwidth]{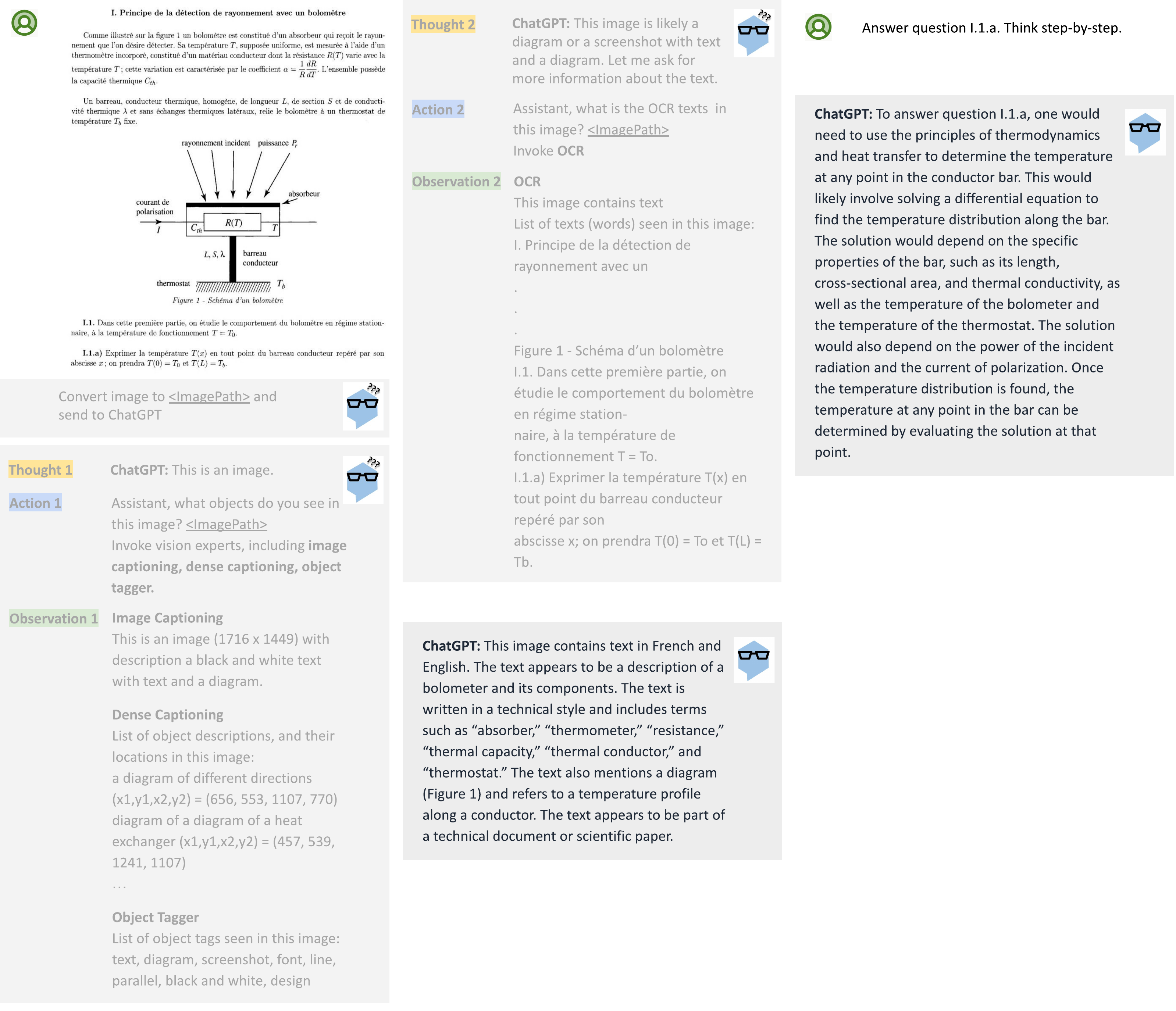}
\end{adjustwidth}
\caption{Unfolded multimodal reasoning and action steps with ChatGPT to tackle physics problem in Figure~\ref{fig:chatgpt_gpt4_1}. 
	}
\label{fig:mmreact-unfolded-physics}
\end{figure*}
%%%%%%%%%%%%%%%%%%%%%%%%%%%%%%%%%%%%%%%%
\section{Conclusion}
We have presented \modelname, a system paradigm that synergizes multimodal reasoning and action to tackle complicated visual understanding problems. \modelname~presents a simple and flexible way to empower LLMs with a pool of vision experts. Extensive zero-shot experiments demonstrate \modelname's capabilities in solving a wide range of challenging understanding tasks, such as multi-image reasoning, multi-hop document understanding, open-world concept understanding, video summarization, and more.

%%%%%%%%% REFERENCES
% {\small
{\small
% {\footnotesize
% \vspace{-1pt}
\subsection*{Acknowledgment}
% \vspace{-2pt}
We would like to express our gratitude to Jianfeng Gao for his valuable suggestions, as well as to Jianwei Yang for generously providing the image editing tool utilizing the X-Decoder framework. 

\bibliographystyle{ieee_fullname}
\bibliography{egbib}
}

\end{document}